\documentclass{article}
\pdfoutput=1

     \PassOptionsToPackage{numbers, compress}{natbib}

\usepackage{arxiv}

\usepackage[utf8]{inputenc} 
\usepackage[T1]{fontenc}    
\usepackage{hyperref}       
\usepackage{url}            
\usepackage{booktabs}       
\usepackage{amsfonts}       
\usepackage{nicefrac}       
\usepackage{microtype}      
\usepackage{xcolor}         
\usepackage{amssymb}
\usepackage{amsmath}
\usepackage{bm}

\usepackage{natbib} 
\usepackage{doi}
\usepackage{cancel}

\usepackage[pdftex]{graphicx}
\usepackage{wrapfig}
\definecolor{pltgreen}{RGB}{87, 182, 147}
\definecolor{pltorange}{RGB}{247, 120, 80}
\definecolor{pltpink}{RGB}{221, 114, 182}
\definecolor{pltpurple}{RGB}{123, 142, 191}

\usepackage[symbol]{footmisc}

\title{Shallow and Deep Nonparametric Convolutions for Gaussian Processes}

\author{
  Thomas M.~McDonald \stepcounter{footnote}\thanks{Equal contribution.} \\
  Department of Computer Science\\
  University of Manchester\\
  \url{thomas.mcdonald-2@postgrad.manchester.ac.uk} \\
   \And
   Magnus Ross \footnotemark[2] \\
   Department of Computer Science \\
   University of Manchester \\
   \url{magnus.ross@postgrad.manchester.ac.uk} \\
   \AND
   Michael T.~Smith \\
   Department of Computer Science \\
   University of Sheffield \\
   \url{m.t.smith@sheffield.ac.uk} \\
   \And
   Mauricio A.~\'Alvarez \\
   Department of Computer Science \\
   University of Manchester \\
   \url{mauricio.alvarezlopez@manchester.ac.uk} \\
}

\date{}
\renewcommand{\headeright}{}

\begin{document}

\maketitle

\begin{abstract}
A key challenge in the practical application of Gaussian processes (GPs) is selecting a proper covariance function. The moving average, or process convolutions, construction of GPs allows some additional flexibility, but still requires choosing a proper smoothing kernel, which is non-trivial. Previous approaches have built covariance functions by using GP priors over the smoothing kernel, and by extension the covariance, as a way to bypass the need to specify it in advance. However, such models have been limited in several ways: they are restricted to single dimensional inputs, e.g. time; they only allow modelling of single outputs and they do not scale to large datasets since inference is not straightforward. In this paper, we introduce a nonparametric process convolution formulation for GPs that alleviates these weaknesses by using a functional sampling approach based on Matheron's rule to perform fast sampling using interdomain inducing variables. Furthermore, we propose a composition of these nonparametric convolutions that serves as an alternative to classic deep GP models, and allows the covariance functions of the intermediate layers to be inferred from the data. We test the performance of our model on benchmarks for single output GPs, multiple output GPs and deep GPs and find that our approach can provide improvements over standard GP models, particularly for larger
datasets.
\end{abstract}

\section{Introduction} \label{sec:intro}

Gaussian processes (GPs) are a widely used method for probabilistic machine learning, that have been applied successfully in many areas \citep{shafieloo2012gaussian, kong2018gaussian, richardson2018gaussian}. A central problem when modelling data with GPs is the choice of a covariance function. The covariance function controls the properties of the functions that the GP places high probability over, therefore selection of an appropriate covariance is crucially important to achieving success when modelling with GPs. When working with a single dimensional input, most commonly in the time series setting, practitioners can inspect the data to determine patterns such as periodicity, long term trends and so on, and construct an appropriate covariance by combining simpler covariance functions that account for these patterns. However this procedure becomes very difficult in higher dimensions, where it is not easy to determine which covariances should be used by simply inspecting the data. Because of this, in high dimensions, practitioners typically revert to using simple covariances, most commonly the exponentiated quadratic (EQ) or Mat\'ern class of kernels. The difficulty of covariance design in high dimensions means that the ability of GPs to represent rich structures present in the data via the covariance function is often not fully utilised. In this work, we present a model that can be applied to problems with both multiple inputs, multiple outputs (or tasks), and can infer the form of the covariance in a nonparametric fashion.

In order to build such a model, we employ the framework of process convolutions (PCs) \citep{higdon2002space, alvarez2012kernels}, in which a base Gaussian process is convolved with a smoothing kernel to generate another Gaussian process with a modified covariance.  The PC framework can be leveraged to infer covariances nonparametrically, by placing a GP prior over the smoothing kernel. \citet{tobar2015learning} introduced the Gaussian process convolution model (GPCM), which uses this mechanism to construct a GP with a nonparametric covariance for data with a single input and output dimension. Recently, \citet{ross2021learning} extended the GPCM to nonlinear process convolutions, and applied the model to problems in systems identification. In this work, we extend and generalise the GPCM to both multiple input and output dimensions, and provide a scalable inference scheme, which additionally allows layers of GPs with nonparametric covariances to be arbitrarily composed to form deep GP models \citep{damianou2013deep, blomqvist2019deep, salimbeni2017doubly}.

\begin{wrapfigure}{r}{0.45\textwidth}
\vspace{-4mm}
\centering
        \includegraphics[width=0.44\textwidth]{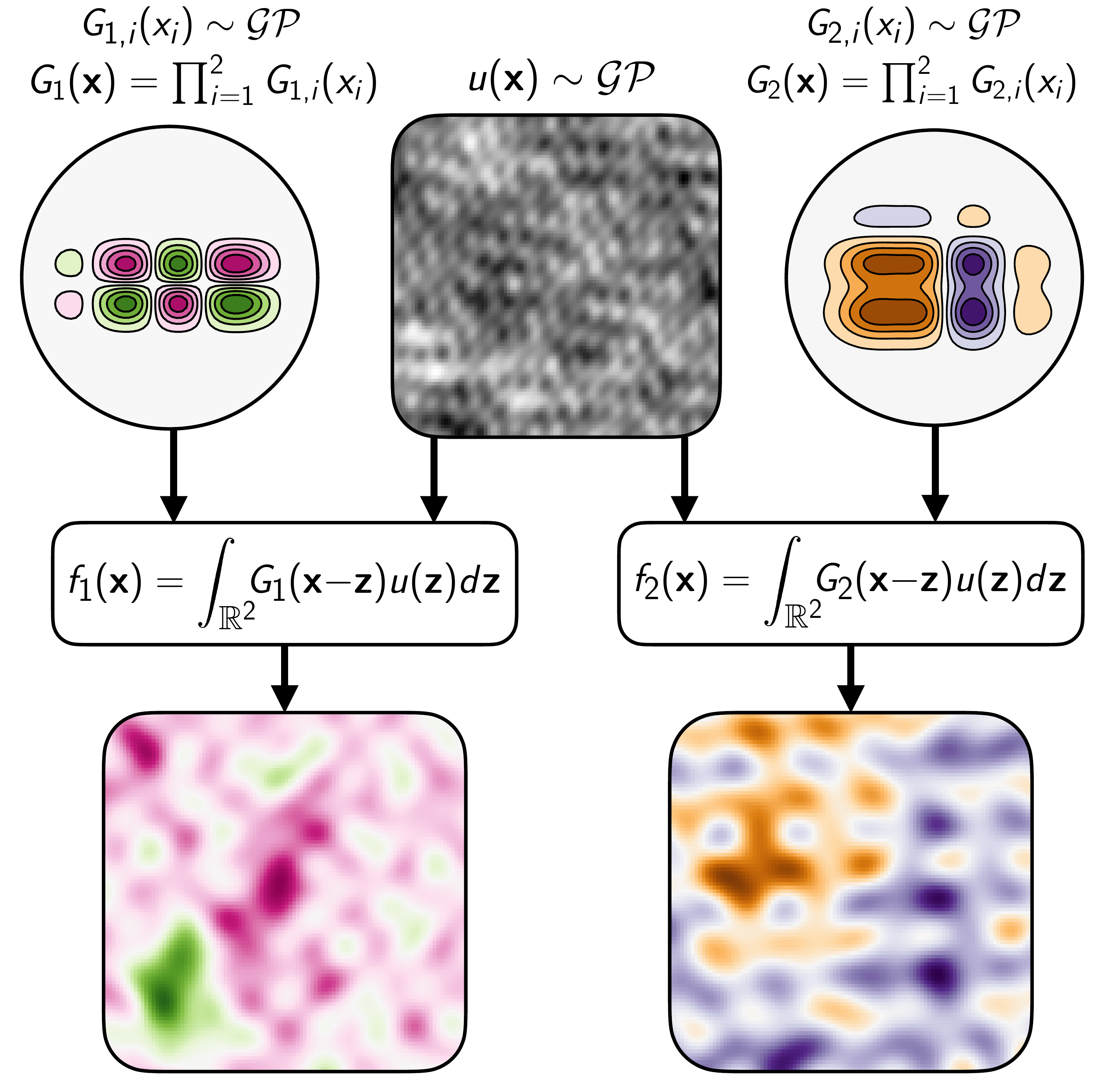}
        \vspace{-2mm}
    \caption{The proposed model.}
    \vspace{-3mm}
    \label{fig:figure1}
\end{wrapfigure}

Figure \ref{fig:figure1} illustrates the proposed generative model for the case of two input dimensions and two outputs. A shared latent function is sampled from a GP with a short lengthscale, which is then smoothed by two distinct convolutional kernels which are generated by a product of GPs, separable over the input dimension, to produce two correlated GP outputs. These distinct convolutional kernels induce different covariance properties in each output, whilst correlations between the outputs across the domain are captured by the shared latent input function. We refer to our method as the nonparametric convolved Gaussian process (NP-CGP). It can be applied with an arbitrary number of inputs, outputs and latent functions.

Many approaches have been proposed to allow for inference of expressive covariance functions. Besides the PC framework, a fruitful avenue has been the design of covariances in the spectral domain. Spectral mixture kernels \citep{wilson2013gaussian, wilson2014fast, jang2017scalable} model the covariance as a mixture of Gaussians in frequency space, and have been extended to multiple outputs \citep{parra2017spectral}, and nonstationary processes \citep{altamirano2022nonstationary, shen2019harmonizable}. Additionally, nonparametric GP priors can also be placed on the spectral density \citep{benton2019function}. Expressive nonstationary covariances can also be constructed by warping the input space with deep neural networks \citep{wilson2016deep}, by the application of stochastic differential equations \cite{pmlr-v89-hegde19a}, or, as in the case of deep GPs \citep{damianou2013deep, blomqvist2019deep, salimbeni2017doubly}, by warping the space with a composition of GPs. The layers of deep GP models are a prime example of a scenario in which we do not necessarily have well-informed prior knowledge regarding the form of the covariance, and as such, learning the covariance directly could be beneficial. We investigate this idea by composing layers of NP-CGPs to form a deep GP model in which the covariances of each layer are themselves inferred from data. 

In this work we present the following four contributions: \textbf{1)} A generalisation of the GPCM to the case of multi-dimensional inputs and outputs with a scalable inference scheme to allow for the use of large datasets. \textbf{2)} An extension to the functional sampling method of \cite{wilson2020efficiently} to cases where the inducing points lie in a transformed space relative to the samples. \textbf{3)} A fast sampling procedure for the model in high dimensions based on the exact integration of the aforementioned approximate functional samples. \textbf{4} A deep GP model for which the covariances of the layers are inferred from the data, formed from a composition of NP-CGP layers. Implementations of the models presented in this work are available at \url{https://github.com/tomcdonald/np-dgp}.

\section{Background} \label{sec:background}
This section briefly reviews the theory behind process convolutions, the GPCM, sampling with pathwise updates and deep GPs.
\subsection{Process convolutions} \label{sec:pc}
The PC framework allows expressive covariances for GPs to be constructed, and can be used to automatically learn the form of covariances from data. In the PC framework \citep{Barry:blackbox96,higdon2002space,alvarez2012kernels}, the function which we wish to model is assumed to have been generated by applying some linear convolution operator to a latent function $u$ represented by a GP, which outputs a new GP \citep{gpmlbook}.  The PC framework can be used to construct multiple output GPs (MOGPs), which allow for inference over vector functions $\mathbf{f}:\mathbb{R}^P \mapsto \mathbb{R}^D$, where $P$ is the number of input dimensions and $D$ is the number of output dimensions. This is useful when we have a set of outputs, represented by the elements of $\mathbf{f}$, which we know are correlated in some way, but also exhibit independent variation. We can construct MOGPs using PCs by assuming each output, $f_d$ is generated by convolving an independent convolutional kernel $G_d$ with a shared latent process $u$, so $f_d(\mathbf{x}) = \int_{\mathcal{X}} G_d(\mathbf{x}-\bm{\tau})u(\bm{\tau})d\bm{\tau}$, where $\mathcal{X}$ is the domain of integration. For many applications it is often overly restrictive to assume that the shared variations can be encapsulated by a single function, and so we instead can use a set of functions $\mathbf{u}:\mathbb{R}^P \mapsto \mathbb{R}^Q$, with each being transformed in a different way for each output. We can express this as $f_d(\mathbf{x}) = \sum^Q_{q=1} \int_{\mathcal{X}} G_{d,q}(\mathbf{x}-\boldsymbol{\tau})u_q(\boldsymbol{\tau})d\boldsymbol{\tau}$. This is the most general form of the model and can be written in the more succinct form, 
\begin{equation}\label{eq:vecPCs}
\begin{split}
\mathbf{f}(\mathbf{x}) &= \int_{\mathcal{X}} \mathbf{G}(\mathbf{x}-\mathbf{z})\mathbf{u}(\mathbf{z})d\mathbf{z} \implies\\ \mathbf{k_f}(\mathbf{x}, \mathbf{x}')&= \int_{\mathcal{X}} \mathbf{G}(\mathbf{x}-\bm{\tau})\mathbf{k_u}(\bm{\tau}, \bm{\tau}')\mathbf{G}^{\top}(\mathbf{x}'-\bm{\tau}')d\bm{\tau} d\bm{\tau}',
\end{split}
\end{equation}
where $\mathbf{G}: \mathbb{R}^P \mapsto \mathbb{R}^{D\times Q}$ consists of square integrable elements, $\mathbf{k_f}: \mathbb{R}^P \times \mathbb{R}^P \mapsto \mathbb{R}^{D\times D}$ is the matrix-valued covariance for the output, and $\mathbf{k_u}: \mathbb{R}^P \times \mathbb{R}^P \mapsto \mathbb{R}^{Q\times Q}$ is the matrix-valued covariance for the shared inputs, which is diagonal due to the assumed independence of the functions representing the elements of $\mathbf{u}$. Eq. \eqref{eq:vecPCs} produces functions with stationary covariance; a nonstationary covariance can be obtained by using a convolutional kernel that varies over the input domain, but this will not be considered in the present work. Various properties of interest can be embedded in $\mathbf{f}$ via $\mathbf{G}$, for example the properties of different physical systems, by using the Green's function of a differential operator \cite{alvarez2009latent}.

\subsection{Gaussian process convolution models} 
The GPCM uses the form of Eq. \eqref{eq:vecPCs} in the case $P,D,Q=1$, and places a GP prior over the convolutional kernel, which in turn induces a prior over the covariance function of the output. To ensure the output is finite, the authors introduce the decaying square exponential (DSE) covariance, which consists of a regular EQ covariance with an additional window, which ensures that the samples decay to zero away from the origin. For the input process, a white noise covariance is used, with the process being summarised by a set of interdomain inducing points, where the interdomain transform is a Gaussian convolution.
The use of white noise for the input process is motivated by the fact that the lengthscale of the output process is bounded from below by the lengthscale of the input process. The white noise input process informally has a lengthscale of zero, therefore by using it, no restriction is placed on the lengthscale of the output.
Since the white noise has zero lengthscale, it cannot be summarised by a finite number of inducing points, which necessitates the interdomain transform. For inference, the GPCM uses a classical mean field variational inference scheme. \citet{bruinsma2022modelling} introduce a number of improvements to both the model structure and inference in the GPCM, particularly extending the model to non-smooth time series using a causal convolution operator, as well as forming a structured variational inference scheme which drastically improves the accuracy and speed of inference. A generalisation of the GPCM to multiple inputs and outputs has previously been discussed by \citet{bruinsma2016generalised}, who coined the name generalised GPCM (GGPCM) to refer to the model. However inference in the model was never implemented, and as such the model was not applied to any data. \citet{ross2021learning} propose the nonparametric Volterra kernels model (NVKM), an extension of the GPCM to nonlinear convolution operators and multiple outputs, which employs doubly stochastic variational inference (DSVI) for approximate inference, and can be used for systems identification, but does not use the interdomain transform for the input process.

\subsection{Sampling GP functions}\label{sec:sampling} \citet{wilson2020efficiently} present a method based on Matheron's rule \citep{etde_5214736}, which allows for the efficient sampling of approximate functions from the posterior of a GP with a stationary covariance. Sampling functions enables samples from the GP at $N$ locations to be evaluated in $O(N)$ time, as opposed to the $O(N^3)$ of standard GP sampling, a significant improvement for applications which require the evaluation of samples at many locations. An additional benefit of sampling \textit{functions} from GPs is that different operators, including integral and differential operators, can be applied to the samples themselves, allowing for the generation of samples from (possibly) highly complex non-Gaussian processes to be obtained efficiently. This idea was used by \citet{ross2021learning} in the context of the NVKM to generate samples from the output of a nonlinear process convolution. \citet{wilson2020efficiently} present Matheron's rule in the context of samples from a GP posterior given inducing variables (or data) $\mathbf{u}$ as
\begin{equation}\label{eq:wilsonsamps}
    \underbrace{(f \mid \mathbf{u})(\cdot)}_{\text {posterior }} \stackrel{\mathrm{d}}{=} \underbrace{f(\cdot)}_{\text {prior }}+\underbrace{k(\cdot, \mathbf{Z}) \mathbf{K}^{-1}\left(\mathbf{u}-\mathbf{f}\right)}_{\text {update }},
\end{equation}
where $\mathbf{K}$ is the covariance matrix of the inducing variables with inputs $\mathbf{Z}$, and $\mathbf{f}=f(\mathbf{Z})$. This expression shows that a functional sample from the posterior, conditioned on data (or inducing outputs) $\mathbf{u}$, can be decomposed into functional samples from the prior, and an update term which accounts for the residual between the prior sample and the data. The key innovation introduced by \citet{wilson2020efficiently} is that we can represent $f(\cdot)$ using random Fourier features (RFFs) \citep{rahami2007random}. Since only the prior, which uses a stationary covariance, uses RFFs, the pathologies associated with the use of RFFs in the nonstationary posterior can be avoided, while still retaining the computational benefits they provide.

\subsection{Deep Gaussian processes} \label{sec:dgp}
Deep Gaussian processes (DGPs) are a class of hierarchical probabilistic models which are capable of modelling nonlinear and nonstationary functions with long range correlations, which are typically difficult to model with standard, shallow GPs \citep{damianou2013deep}. For a given input $\mathbf{X} \in \mathbb{R}^{N \times P}$, a typical compositional DGP model takes the form,
\begin{equation} \label{eq:dgp}
    \mathbf{F}^L = \mathbf{f}^{L}\circ \mathbf{f}^{L-1} \circ \dots \circ \mathbf{f}^{1}(\mathbf{X}),
\end{equation}
where $\mathbf{f}^\ell$ is used to represent the functions themselves at the $\ell$-th layer, whilst the function \textit{values} are denoted by $\mathbf{F}^\ell = \mathbf{f}^\ell(\mathbf{F}^{\ell - 1})$, and $\mathbf{F}^0=\mathbf{X}$. \citet{salimbeni2017doubly} introduced an unbiased DSVI scheme for DGPs consisting of a variational posterior and evidence lower bound (ELBO) which are not based on the simplifying assumption of independence between layers. Due to its computational efficiency and empirical performance compared to other DGP inference techniques \citep{bui2016deep}, we utilise DSVI in this work to perform approximate Bayesian inference.

\section{Nonparametric convolutions for Gaussian processes}
In this section, we present a generalised PC model of the form shown in Eq. \eqref{eq:vecPCs} which jointly infers vector-valued functions $\mathbf{f}:\mathbb{R}^P \mapsto \mathbb{R}^D$, and their corresponding nonparametric convolutional kernel $\mathbf{G}$. This induces a nonparametric matrix-valued covariance, $\mathbf{k_f}$, for $\mathbf{f}$.
\label{sec:NP-CGP}
 \subsection{Single-dimensional inputs} Before presenting the multi-dimensional version of our model, we start with the single-dimensional input case ($P=1$) in Eq. \eqref{eq:vecPCs}, given as $\mathbf{f}(x) = \int_{\mathbb{R}} \mathbf{G}(x-z)\mathbf{u}(z)dz$, where each entry in the vector $\mathbf{u}(x)$ follows a GP, i.e. $u_q(x) \sim \mathcal{GP}[0, k_{u_q}(x, x')], q=1, \dots,Q$; and each entry in the matrix $\mathbf{G}(x)$ also follows a GP, i.e. $G_{d, q}(x) \sim \mathcal{GP}[0, k_{G_{d,q}}(x, x')],q=1, \dots, Q$ and $d=1, \dots, D$. Throughout this work, we use the DSE covariance, described by \citet{tobar2015learning}, for the elements of $\mathbf{G}$ to ensure that they are square integrable. This model can be seen as a generalisation of the GPCM, which can only represent functions from $\mathbb{R} \mapsto \mathbb{R}$, to the multi-output case. Both exact sampling and exact inference in the NP-CGP model above are intractable as the integral cannot be computed when $G_{d,q}(\cdot)$ and $u_q(\cdot)$ are infinite dimensional stochastic processes. In order to draw samples from the model, we must first summarise the GPs representing the convolutional kernels and the input processes with finite collections of inducing points. These inducing points can then be used to sample approximate functions from the convolutional kernels and input processes, which can be integrated exactly to produce samples from the output. Fast and accurate sampling allows a doubly stochastic variational inference scheme to be constructed for the model, which is discussed later in this section.

\subsection{Multi-dimensional inputs} To extend the model above to the multi-dimensional input case, we need to perform inference on the model $\mathbf{f}(\mathbf{x}) = \int_{\mathbb{R}^P} \mathbf{G}(\mathbf{x}-\mathbf{z})\mathbf{u}(\mathbf{z})d\mathbf{z}$. Following the same construction used for the single-dimensional input case, we place GP priors over the inputs of $\mathbf{u}(\mathbf{x})$, i.e. $u_q(\mathbf{x}) \sim \mathcal{GP}[0, k_{u_q}(\mathbf{x}, \mathbf{x}')], q=1, \dots,Q$. In a similar way, one could place GP priors over the individual elements of $\mathbf{G}$, i.e. $ G_{d, q}(\mathbf{x}) \sim \mathcal{GP}[0, k_{G_{d,q}}(\mathbf{x}, \mathbf{x}')]$. Given that our inference approach is based on inducing points, this option of GP priors for $G_{d, q}(\mathbf{x})$ is computationally intractable in high input dimensions. The number of inducing points required to characterise $G_{d,q}$ increases exponentially with the number of input dimensions, as the inducing points become increasingly sparsely distributed across the input space in higher dimensions. This is a problem for the convolutional kernels in particular because they operate over the whole domain, so any uncertainty in their value translates across the entire output function.

We can address this problem by modelling $G_{d,q}$ as product separable, such that $G_{d,q}(\mathbf{x})=\prod^{P}_{p=1}G^{(p)}_{d,q}(x_p)$, where each $G^{(p)}_{d, q}$ is an independent GP with its own set of inducing points, and $x_p$ is the $p$-th dimension of the input. This assumption allows us to characterise $G_{d,q}$ using a set of inducing points whose size scales linearly with the number of input dimensions, which is a considerable improvement. Furthermore, if we were to assume that each degree of freedom for $G^{(p)}_{d,q}(x_p)$ were to be modelled as a GP, we would need $DQP$ independent GPs
to model all the elements in $\mathbf{G}$. Therefore, to reduce the number of GPs used, we further assume that each smoothing kernel $G_{d,q}(\mathbf{x})$ can be expressed as
$G_{d,q}(\mathbf{x})=a_q G_{d}(\mathbf{x})$, where $a_q\in\mathbb{R}$, reducing the number of  GPs to model $\mathbf{G}$ to $DP$. The generative model is given as
\begin{equation}\label{eq:NP-CGP}
     \begin{split}
u_q(\mathbf{x}) &\sim \mathcal{GP}[0, k_{u_q}(\mathbf{x}, \mathbf{x}')], \\ 
         G^{(p)}_{d}(x_p) &\sim \mathcal{GP}[0, k_{G^{(p)}_{d}}(x_p, x'_p)],\\
         G_{d, q}(\mathbf{x})& = a_q G_{d}(\mathbf{x}), \quad  G_{d}(\mathbf{x}) =\prod_{p=1}^P G^{(p)}_{d}(x_p) ,  \\ 
\mathbf{f}(\mathbf{x}) &= \int_{\mathbb{R}^P} \mathbf{G}(\mathbf{x}-\mathbf{z})\mathbf{u}(\mathbf{z})d\mathbf{z},
       \end{split}
\end{equation}
with $q ={1, \dots, Q}$, $p = {1, \dots, P}$ and $d= {1, \dots, D}$.
We refer to the model in Eq. \eqref{eq:NP-CGP} as the nonparametric convolved Gaussian process (NP-CGP). Notice that $G_{d,q}(\mathbf{x})$ is not a GP, though the output process conditioned on $G_{d,q}(\mathbf{x})$ will still be a GP as the convolution is a linear operator on the input process. The separable restriction on $G_{d,q}(\mathbf{x})$ corresponds to the restriction that the covariance function is a product of covariances for each input, which is true for most popular multi-dimensional covariances, such as the \textit{automatic relevance determination} (ARD) kernel. Whilst specifying a product separable convolutional kernel results in a computationally feasible form of the model, we also present a more efficient variant, which we refer to as the \textit{Fast NP-CGP} (FNP-CGP). In this case, rather than using separate convolutional kernels $G^{(p)}_{d}$ per input and output dimension (resulting in a total of $DP$ kernels), we share a single convolutional kernel across $d=1,...,D$, with each being convolved with a different linear combination of input functions, i.e. $G_{d, q}(\mathbf{x})= a_{d, q} \prod_{p=1}^P G^{(p)}(x_p)$.

\subsection{Interdomain input processes} 
\begin{wrapfigure}{r}{0.35\textwidth}
\vspace{-6mm}
\centering
        \includegraphics[width=0.34\textwidth]{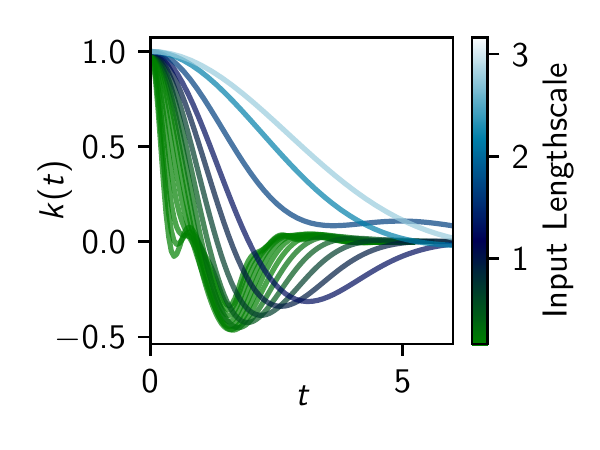}
        \vspace{-5mm}
    \caption{Samples of output covariance, shaded by input process lengthscale.}
    \vspace{-4mm}
    \label{fig:cov_intpolation}
\end{wrapfigure}
When considering this model, it may seem as though we are free to choose any covariance we please for the input process, and place inducing points in the same space. Indeed, \citet{ross2021learning} take this approach for a model with a similar structure. However, for the case of high dimensional inputs, this approach is no longer suitable. As mentioned previously, in high dimensions it is not computationally feasible to use a number of inducing points that will densely cover the space, and as such the lengthscale of the process the inducing points are placed upon must become large. In PC models that use a smoothing transform, such as the GPCM, the lengthscale of the output is, in a sense, bounded from below by the lengthscale of the input process, as the smoothing transform can never increase its lengthscale. This presents an issue for the NP-CGP, because the necessity of a long lengthscale input process in high dimensions would make it difficult to learn expressive covariances. Figure \ref{fig:cov_intpolation} shows a covariance sample from the output smoothly interpolated between a variety of input lengthscales, with all else remaining the same. We can see that as the input lengthscale increases, the complex structures in the output process covariance disappear, and the covariance tends to that of the input. One way to address this problem is to use the framework of interdomain inducing points \citep{lazaro2009inter, alvarez2010efficient}, using a short lengthscale for the input process to allow the covariances of the output to remain expressive, and placing inducing points within a smoothed domain. This allows us to retain the ability to summarise the variation of the process over long lengthscales, as is necessary in high dimensions.

\subsection{Sampling from the outputs}
Inference in the NP-CGP is predicated upon the ability to generate samples from the model outputs efficiently. To achieve this, we utilise the method of \citet{wilson2020efficiently} to produce function samples from the input process, $\mathbf{u}^{(s)}(\mathbf{x})$ and convolutional kernel process $\mathbf{G}^{(s)}(\mathbf{x})$, which we then analytically integrate through the convolution integral, to obtain the output as $\mathbf{f}^{(s)}(\mathbf{x}) = \int_{\mathbb{R}^P} \mathbf{G}^{(s)}(\mathbf{x}-\mathbf{z})\mathbf{u}^{(s)}(\mathbf{z})d\mathbf{z}$. Although the computation is somewhat involved, closed form solutions to this integral can be obtained. Further details regarding this computation are available in the supplemental material.

\subsection{Sampling interdomain functions}
As discussed above, it is necessary to use the framework of interdomain inducing points for the functions $u_q$, to maintain the ability to represent expressive covariances in high dimensions. Sampling and inference in our model also requires access to functional samples for the functions $u_q$, and as such we must combine the two methods. We represent the elements of our input process using an EQ covariance with a short lengthscale, with the interdomain process $\tilde{\mathbf{u}}$ being generated by a smoothing transformation with a Gaussian window, $\tilde{u}_q(\mathbf{x}) = \int_{\mathbb{R}^P} g_q(\mathbf{z}, \mathbf{x}) u_q(\mathbf{z}) d\mathbf{z}$, with $g_q(\mathbf{z}, \mathbf{x}) = a_q e^{-\sum_p \alpha_{q,p}(x_p-z_p)^2}$, where $\alpha_{q,p}$ is related to the lengthscale of the transformation. Matheron's rule, which the method of \citet{wilson2020efficiently} relies upon, applies for collections of jointly distributed Gaussian variables, and as such, can be readily adapted to the interdomain case, because the processes $u_q$ and $\tilde{u}_q$ are jointly Gaussian. The expression for a functional sample from $u_q$ in the interdomain case now becomes 
\begin{equation}\label{eq:npdgp_layer_samps}
\begin{split}
    u_q^{(s)}(\cdot) = &\sum^{B}_{i=1}w_i\phi_i(\cdot) \\ &+ k_{u_q, \tilde u_q}(\cdot, \mathbf{z}^{\tilde u_{q}}) \mathbf{K}_{\tilde u_q, \tilde u_q}^{-1}\left(\mathbf{v}^{\tilde u_{q}}-\boldsymbol{\tilde\Phi}\boldsymbol{w}\right),
\end{split}
\end{equation}
where $\phi_i$ is one of $B$ RFF basis functions with random weights $w_i\sim\mathcal{N}(0, 1)$, $k_{u_{q}, \tilde u_{q}}$ represents the cross-covariance between domains, $\mathbf{z}^{\tilde u_{q}} \in \mathbb{R}^{M_u \times P}$ is the set of $M_u$ inducing inputs with corresponding outputs $\mathbf{v}^{\tilde u_{q}} \in \mathbb{R}^{M_u \times 1}$, $\mathbf{K}_{\tilde u_{q}, \tilde u_{q}}$ is the covariance of the inducing points in the transformed domain, and
$\boldsymbol{\tilde\Phi}\in \mathbb{R}^{M_u \times B}$ is a matrix with each of the basis functions evaluated in the transformed domain for each inducing input. These transformed basis functions can be computed by applying the interdomain convolution, such that $\tilde\phi_i(\mathbf{x}) = \int_{\mathbb{R}^P} g_q(\mathbf{z}, \mathbf{x}) \phi_i(\mathbf{z}) d\mathbf{z}$. For the Gaussian transform discussed above, we can obtain expressions for transformed basis functions in closed form, with details of the computation included in the supplemental material, alongside further information regarding the RFF basis, the derivation of \eqref{eq:npdgp_layer_samps} and the computation of the various interdomain and cross-covariances. To the best of our knowledge, this method for fast sampling of interdomain GPs is yet to appear in the literature. 

\subsection{Doubly stochastic variational inference}
Following the approach of \citet{salimbeni2017doubly}, we employ DSVI to perform approximate inference in the the NP-CGP and FNP-CGP. Firstly, we must introduce the inducing points for the convolutional kernel processes, $\mathbf{v}^{G_{d,q}} \in \mathbb{R}^{M_G\times 1}$, with entries $v^{G_{d,q}}_i = G_{d,q}(z^{G_{d,q}}_i), i=1, \dots, M_G$, where $M_G$ is the number of inducing points used. The associated inducing inputs are denoted as $z^{G_{d,q}}_i, i=1, \dots, M_G$, which we collect into $\mathbf{z}^{G_{d, q}}\in \mathbb{R}^{M_G \times 1}$. Additionally, to simplify the notation, we collect all of the inducing points for the convolutional kernel and input processes into $\mathbf{V}^{G}=\{\mathbf{v}^{G_{d, q}}\}^{D, Q}_{d, q=1}$ and $\mathbf{V}^{\tilde{u}}=\{\mathbf{v}^{\tilde{u}_{q}}\}^{Q}_{ q=1}$ respectively. If we consider some input data $\mathbf{X}\in\mathbb{R}^{N\times P}$ with corresponding outputs $\mathbf{Y}\in\mathbb{R}^{N\times D}$, we can express the joint distribution of the NP-CGP as,
\begin{equation} \label{NP-CGP_joint}
\begin{split}
    p(\mathbf{Y}, \mathbf{G}, \mathbf{V}^{\mathbf{G}} , &\mathbf{u}, \mathbf{V}^{\tilde{\mathbf{u}}}) =\prod^{N}_{i=1}  p(\mathbf{y}_{i}|\mathbf{F}_i)\\
&\times p(\mathbf{G} | \mathbf{V}^{\mathbf{G}})p(\mathbf{V}^{\mathbf{G}})p(\mathbf{u} | \mathbf{V}^{\tilde{\mathbf{u}}}) p(\mathbf{V}^{\tilde{\mathbf{u}}}) ,
\end{split}
\end{equation}
where the likelihood is given by $p(\mathbf{y}_{i}|\mathbf{F}_i) = \mathcal{N}(\mathbf{y}_i ; \mathbf{F}_i, \sigma^2_Y)$ and $\mathbf{F}_i = \mathbf{f}(\mathbf{X}_i)$ represents the output of the model for the $i$-th input. As all of the convolutional kernel and input GPs are independent, we have $p(\mathbf{G} | \mathbf{V}^{\mathbf{G}}) = \prod^{D}_{d=1} p( G_{d}| \mathbf{v}^{G_{d}})$  where $p( G_{d}| \mathbf{v}^{G_{d}})$ is the GP posterior distribution given the inducing points, and likewise $p(\mathbf{u} | \mathbf{V}^{\tilde{\mathbf{u}}}) = \prod^{Q}_{q=1} p(u_{q}| \mathbf{v}^{\tilde{u}_{q}})$, where again $p(u_{q}| \mathbf{v}^{\tilde{u}_{q}})$ are GP posteriors. $p(\mathbf{V}^{\mathbf{G}})$ and $p(\mathbf{V}^{\tilde{\mathbf{u}}})$ represent the priors over the inducing points. Following the approach of \citet{tobar2015learning}, we employ a mean-field variational posterior, which takes the form,
\begin{equation} \label{NP-CGP_posterior}
    q(\mathbf{G}, \mathbf{V}^{\mathbf{G}} , \mathbf{u}, \mathbf{V}^{\tilde{\mathbf{u}}}) = p(\mathbf{G} | \mathbf{V}^{\mathbf{G}})q(\mathbf{V}^{\mathbf{G}})p(\mathbf{u} | \mathbf{V}^{\tilde{\mathbf{u}}})q(\mathbf{V}^{\tilde{\mathbf{u}}}),
\end{equation}
where $q(\mathbf{V}^{\mathbf{G}}) =\prod^{D}_{d=1} \mathcal{N}(\mathbf{v}^{G_{d}};\boldsymbol{\mu}^{G_{d}}, \boldsymbol{\Sigma}^{G_{d}})$  and $q(\mathbf{V}^{\tilde{\mathbf{u}}}) =\prod^{Q}_{ q=1} \mathcal{N}(\mathbf{v}^{\tilde{u}_{q}};\boldsymbol{\mu}^{\tilde{u}_{q}}, \boldsymbol{\Sigma}^{\tilde{u}_{q}})$ are variational distributions, whose means and covariance matrices are variational parameters. We use samples from both of these variational distributions in the process of analytically computing the functional samples from our model. For ease of exposition, we have omitted the factorisation of the posterior over the input dimensionality. Using $s=1,...,S$ samples from the model, denoted by $\mathbf{F}_i^{(s)} = \mathbf{f}^{(s)}(\mathbf{X}_i)$, and representing the KL divergence as $\text{KL}[\cdot\lVert\cdot]$, we can approximate the variational lower bound as,
\begin{equation} \label{eq:NP-CGP_elbo_main}
\begin{split}
    \mathcal{L} =& \frac{1}{S} \sum^S_{s=1} \log p(\mathbf{y}_i | \mathbf{F}_i^{(s)})\\ &-  \text{KL}[q(\mathbf{V}^{\tilde{\mathbf{u}}}) \lVert p(\mathbf{V}^{\tilde{\mathbf{u}}})] - \text{KL}[q(\mathbf{V}^{\mathbf{G}}) \lVert p(\mathbf{V}^{\mathbf{G}})].
\end{split}
\end{equation}
An extended derivation of the bound above is provided in the supplemental material.

\subsection{Deep convolutions}
As discussed in Section \ref{sec:intro}, one of the key motivations for learning covariances directly from data is that in certain scenarios, such as the case of high-dimensional inputs, it is challenging to specify an appropriate form for the covariance. DGPs are another such example of a situation in which we cannot intuitively reason about the ideal form for the covariance due to the compositional nature of such models; this is especially true within the internal layers of a DGP. As the FNP-CGP is capable of modelling multiple inputs and outputs efficiently using only $P$ convolutional kernels, we can form a compositional DGP with nonparametric covariances from a series of FNP-CGP layers, which we term the \textit{NP-DGP}. The inference procedure for the NP-DGP broadly follows the same process as presented for the FNP-CGP in Section \ref{sec:NP-CGP}, but with a variational posterior which is now factorised across the $L$ layers of the model, 
\begin{equation} \label{dgpnp_posterior_main}
    q(\{\mathbf{G}^\ell, \mathbf{V}^{\mathbf{G}^\ell} , \mathbf{u}^\ell, \mathbf{V}^{\tilde{\mathbf{u}}^\ell} \}^L_{\ell = 1}) = \prod^L_{\ell = 1} p(\mathbf{G}^\ell | \mathbf{V}^{\mathbf{G}^\ell})q(\mathbf{V}^{\mathbf{G}^\ell})p(\mathbf{u}^\ell | \mathbf{V}^{\tilde{\mathbf{u}}^\ell})q(\mathbf{V}^{\tilde{\mathbf{u}}^\ell}).
\end{equation}
This leads to the following approximation to the doubly stochastic variational lower bound,
\begin{align} \label{dgpnp_elbo_main}
    \mathcal{L} = \frac{1}{S} \sum^S_{s=1} \log p(\mathbf{y}_i | \mathbf{F}_i^{L^{(s)}}) - \sum^L_{\ell = 1} \left[ \text{KL}[q(\mathbf{V}^{\tilde{\mathbf{u}}^\ell}) \lVert p(\mathbf{V}^{\tilde{\mathbf{u}}^\ell})] + \text{KL}[q(\mathbf{V}^{\mathbf{G}^\ell}) \lVert p(\mathbf{V}^{\mathbf{G}^\ell})] \right].
\end{align}
A full derivation of this bound is included in the supplemental material; for the case where $L=1$, this derivation also serves as a full derivation of the NP-CGP bound shown in Eq. \ref{eq:NP-CGP_elbo_main}. Also included in the supplemental material are the the pathwise sampling expressions which allow us to analytically map samples from the convolutional and input processes at each layer through Eq. \eqref{eq:NP-CGP}. 
 
\section{Related Work}
\citet{ross2021learning} propose the \textit{nonparametric Volterra kernels model} (NVKM), an extension of the GPCM to nonlinear systems and multiple outputs, which employs DSVI for approximate inference. The NP-CGP can be seen as extension of this model to multiple input dimensions, however unlike this work and that of \citet{tobar2015learning}, the authors do not use interdomain inducing points for the $u$ process. Additionally, unlike \citet{tobar2015learning}, the authors use an EQ covariance for the $u$ process, rather than a Dirac delta covariance. The NVKM does exploit the same efficient sampling scheme as our model, based on pathwise updates, as discussed in Section \ref{sec:sampling}.

\textit{Spectral mixture kernels} are an alternative approach to automatic learning of covariance functions from data, which involves modelling the power spectral density (PSD) of a kernel with a Gaussian mixture and taking the inverse Fourier transform of the PSD to obtain the covariance \citep{wilson2013gaussian}. Building on this work, \citet{parra2017spectral} present the \textit{multi-output spectral mixture} (MOSM), which involves using a multivariate extension to Bochner's theorem in order to extend this approach to MOGPs. A later work by \citet{altamirano2022nonstationary} further extends the MOSM to the case of nonstationary data using harmonizable kernels which automatically identify nonstationary behaviour. \citet{benton2019function} present another means of nonparametric learning of covariances which involves representing the log of the PSD with a GP and applying Bochner's theorem to yield a covariance function. This approach allows for exact GP inference after the covariance has been approximated, however as a result the model is not as scalable as the NP-CGP.

In this work, we propose a novel approach to incorporating convolutions into a DGP, although there has been some prior work in this area. The \textit{deep latent force model} (DLFM) was introduced by \citet{mcdonald2021compositional} as a means of incorporating GPs based on the PC framework into a DGP, which is similar in scope to our work. However, in this case, rather than working with a nonparametric covariance function, the authors derive random Fourier features corresponding to a parametric form for $G$ which is the Green's function of a first order ordinary differential equation. These random features are then integrated these into a multi-layered DGP architecture. \citet{shen2020learning} present an alternative approach to incorporating an expressive kernel into a DGP, deriving the \textit{convolutional spectral kernel} (CSK), a nonstationary, expressive kernel formed from the convolution of two imaginary radial basis functions. However, whilst our deep model is a recursive composition of functions of the form shown in Eq. \eqref{eq:dgp}, known as a \textit{compositional DGP}, \citet{shen2020learning} incorporate the CSK into a \textit{covariance function DGP} which consists of a recursive composition of kernels. \citet{dunlop2018deep} present a comprehensive overview of the relationships between these (and other) DGP formulations. Finally, a key point to note is that our model is unrelated to the \textit{convolutional DGP} of \citet{blomqvist2019deep}, as the convolutions used there are discrete and are applied to the problem of computer vision.

\section{Experiments} \label{sec:experiments}
\subsection{Toy experiment}
\begin{figure}[t]
    \centering
    \includegraphics[width=1.0\textwidth]{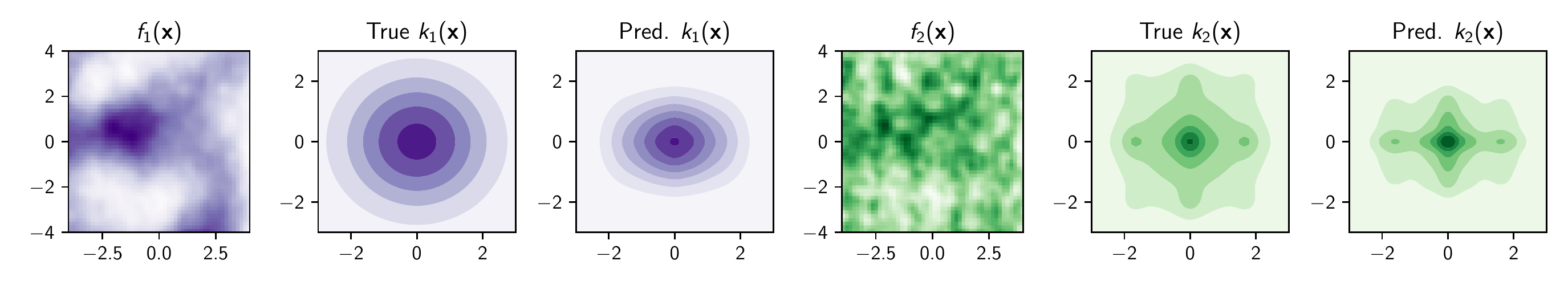}
    \caption{Inferred covariance for each input dimension and output for the toy experiment, with true functions for each output}
    \vspace{-6mm}
    \label{fig:toy_exp}
\end{figure}
Firstly, we present a toy experiment which shows the ability of our model to learn known covariances in multiple dimensions. We use a two dimensional input $\mathbf{x} = [x_1, x_2]^\top$ and generate the ground truth outputs $f_1(\mathbf{x})$ and $f_2(\mathbf{x})$ by sampling from two different linear combinations of $u_{EQ} \sim \mathcal{GP}(0, k_{EQ})$ and $u_P \sim \mathcal{GP}(0, k_P)$, which are GP priors with an EQ and weakly periodic kernel respectively. From the results shown in Figure \ref{fig:toy_exp}, we see that the NP-CGP is capable of recovering the form of both of these covariances. We can also see that the model is overconfident in places, likely due to the somewhat restrictive mean field assumption. Further details regarding this experiment, the experiments later in this section, and all of the data used, are provided in the supplemental material.

\subsection{UCI regression} \label{sec:uci}
\begin{figure}[t]
\centering
        \includegraphics[width=0.9\textwidth]{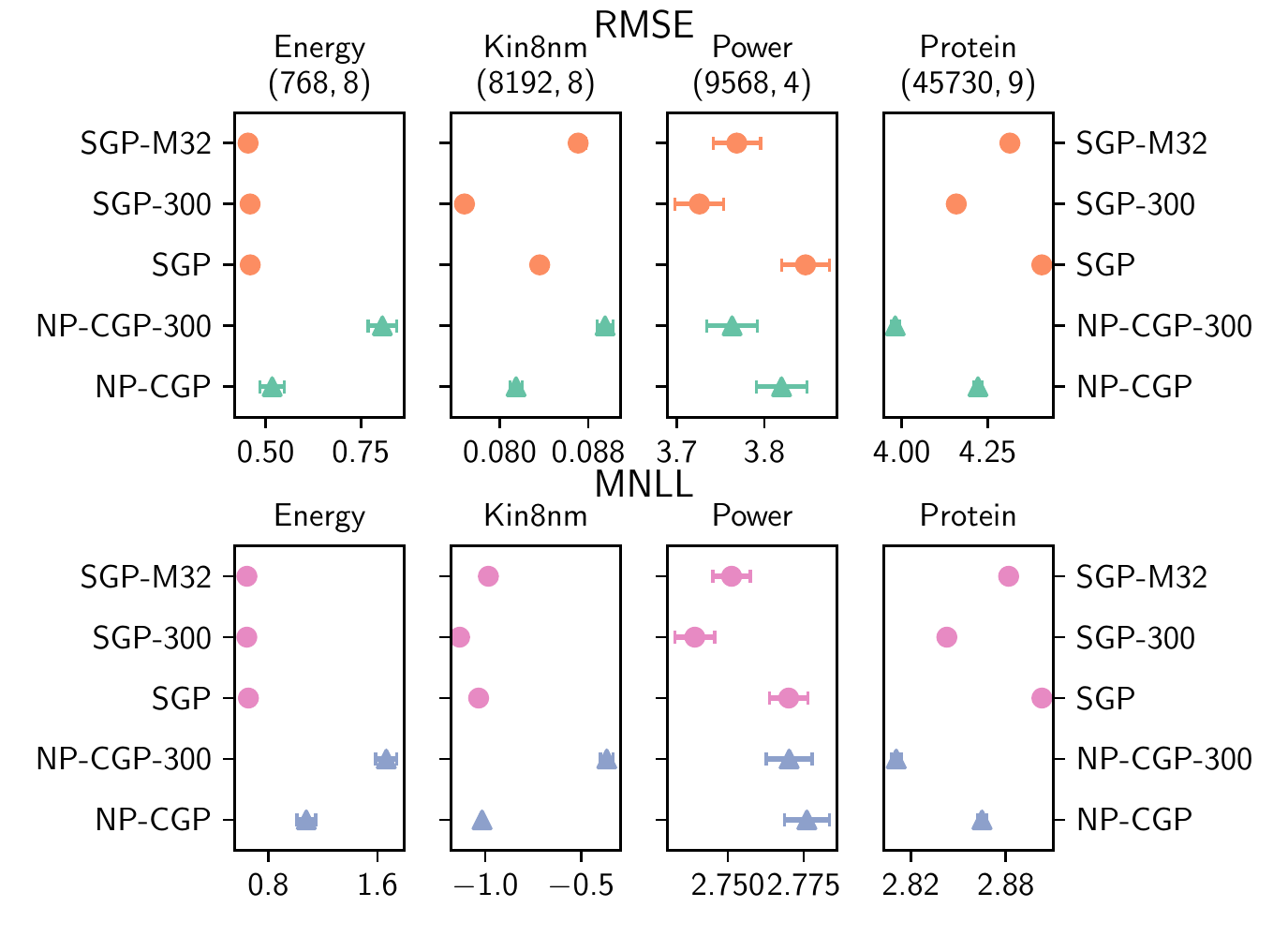}
    \caption{UCI regression results, showing mean and standard error over 20 train/test splits of RMSE and MNLL for our models ($\textcolor{pltgreen}\blacktriangle$ and $\textcolor{pltpurple}\blacktriangle$) and of the baselines ($\textcolor{pltorange}\bullet$ and $\textcolor{pltpink}\bullet$). Lower (i.e. further to the left) is better.}
    \label{fig:shallow_boxplot}
\end{figure}
We evaluate our shallow model on four multi-input, single output UCI regression benchmark datasets \citep{Dua_2019}. We specifically chose datasets which have been shown to benefit from additional model complexity in previous work \citep{salimbeni2017doubly} due their size, complex structure or a combination of the these factors. The results are presented in Figure \ref{fig:shallow_boxplot}, where we compare our shallow model with 100 (NP-CGP) and 300 (NP-CGP-300) inducing points respectively, to shallow stochastic variational GPs with EQ kernels which also use 100 (SGP) and 300 (SGP-300) inducing points. Additionally, we compare to a SGP with 100 inducing points and the Mat\'ern 3/2 kernel (SGP-M32). We find that the NP-CGP with 100 inducing points outperforms or is comparable with all SGP model variants across all metrics and datasets. Adding additional inducing points provides a large performance increase to the NP-CGP for \textit{protein} and \textit{power}, the two largest datasets, but a decrease for the other datasets. These results show that the NP-CGP is a compelling model for single output GP regression tasks.

\begin{figure}[t!]
    \centering
    \includegraphics[width=1.0\textwidth]{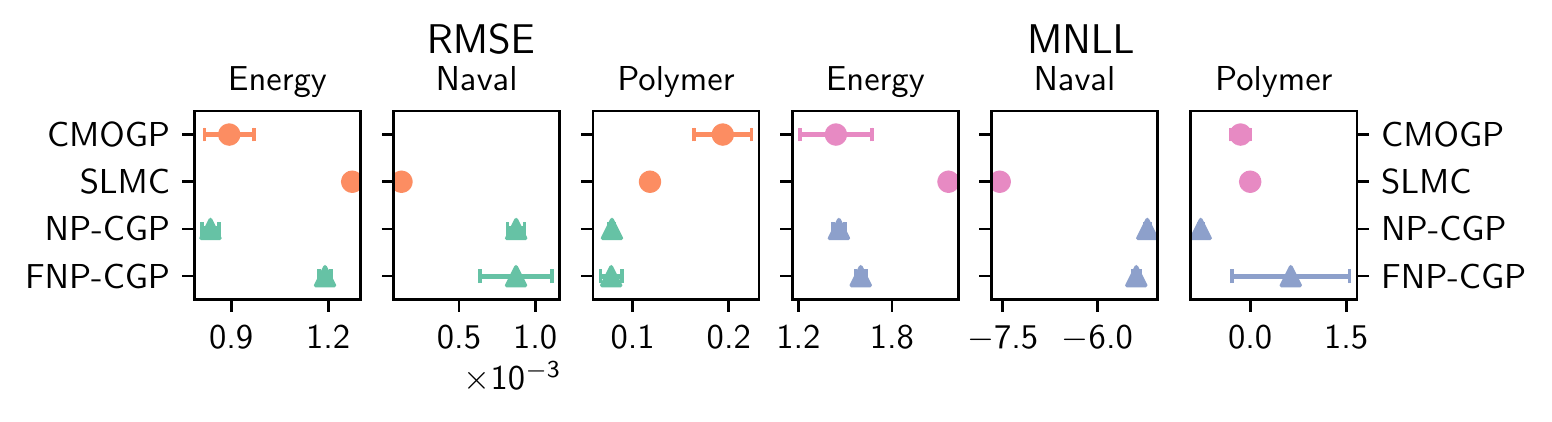}
    \vspace{-5mm}
    \caption{Results over 20 train/test splits for the experiments with multiple inputs and outputs.}
    \label{fig:mo_boxplot}
    \vspace{-5mm}
\end{figure}

\paragraph{Regression with multiple inputs and outputs}
Additionally, to demonstrate the utility of our approach for general regression with multiple inputs and outputs we fit the model on three datasets of that type: \textit{energy} and \textit{naval} medium size datasets with two outputs, and \textit{polymer} a small dataset with four outputs. In Figure \ref{fig:mo_boxplot}, we present results for the full (NP-CGP) and fast (FNP-CGP) variants of our model, alongside a stochastic variational MOGP which uses the \textit{linear model of coregionalization} (SLMC) \citep{alvarez2012kernels} and an exact convolved MOGP (CMOGP)\footnote{No results available for CMOGP on \textit{naval} as dataset is too large for inference in a reasonable amount of time.}\citep{alvarez2011computationally}, which is comparable to our model, but instead with parametric convolutions. We find that our NP-CGP either matches or exceeds the performance of the other models tested for the \textit{energy} and \textit{polymer} datasets, and whilst the SLMC exhibits improved performance for \textit{naval}, however this dataset is known to be easy for GP models to fit well \citep{salimbeni2017doubly}. Additionally, the improvement is only marginal when the small scales of the errors and log-likelihood are taken into account. These results illustrate the benefits of nonparametric convolutions for regression problems with multiple outputs.

\subsection{Large-scale regression}\label{sec:large_scale}
To show the scalability of our model to regression problems with hundreds of thousands of observations, we evaluate the NP-CGP on the \textit{airline} dataset. This is a commonly used large-scale single-output regression benchmark for GPs, where each observation has dimensionality $P=8$. Specifically, we use the first 700k observations for training and the next 100k for testing, with a mini-batch size of 10k.

The results displayed in Table \ref{tab:airline} show that the NP-CGP with 100 inducing points outperforms conventional variational GPs with both 100 and 500 inducing points (SGP and SGP-500). Additionally, we see that the NP-CGP also achieves superior performance to the two layer deep GP of \citet{salimbeni2017doubly} (DGP2). The latter is a particularly encouraging result, as it suggests that the performance gap between shallow and hierarchical models may be bridged by utilising our nonparametric approach to learning covariances, although it should be noted that adding additional layers to the DGP can increase performance further, beyond that of our model.

\begin{figure}
    \centering
    \includegraphics[width=0.9\textwidth]{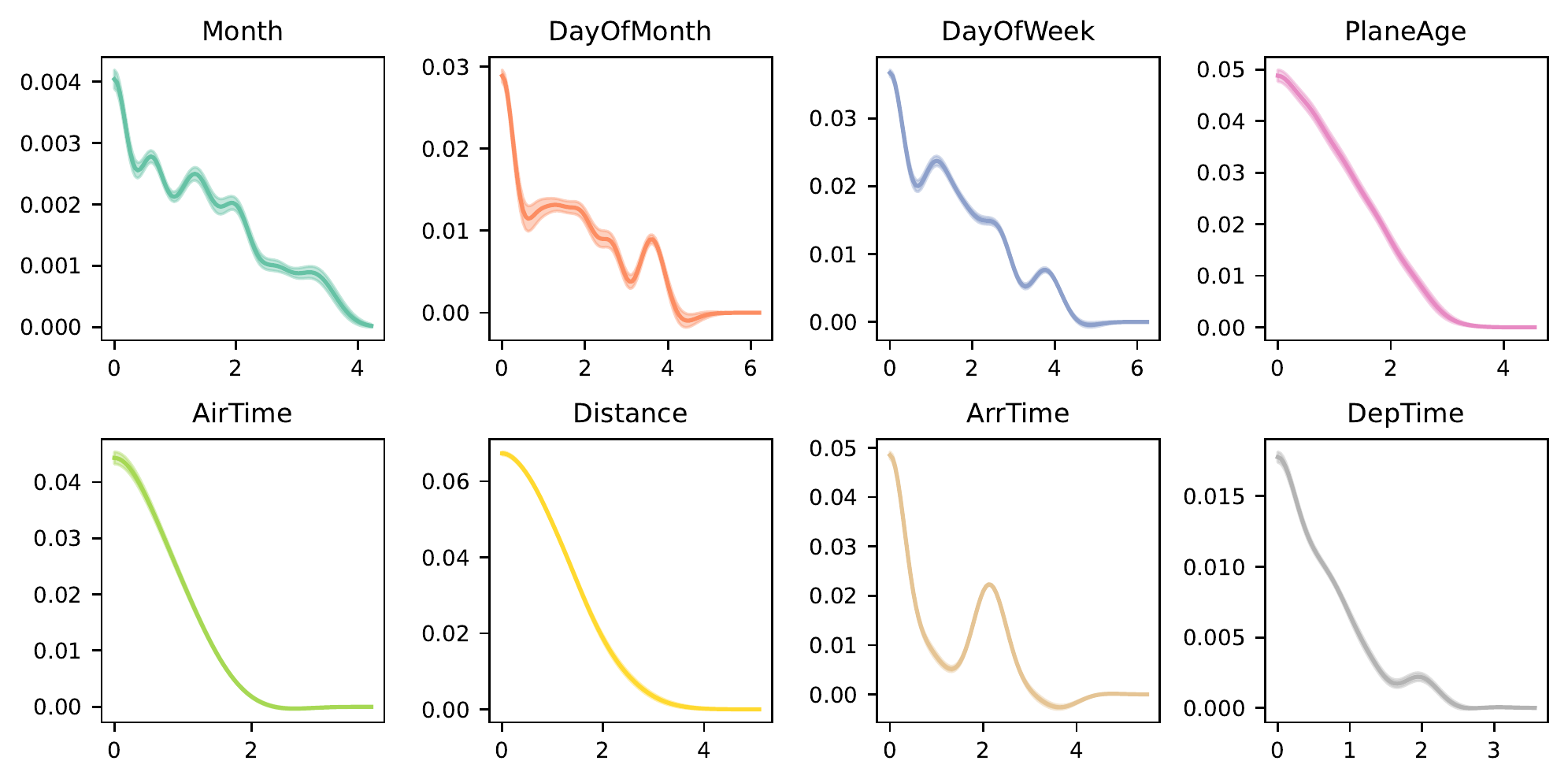}
    \caption{Visualisation of the covariances learned by the NP-CGP for the airline dataset. The solid line is the mean computed using 50 samples from the model, and the shaded confidence interval denotes $\pm 2\sigma$.}
    \label{fig:airline_covariances}
\end{figure}

Figure \ref{fig:airline_covariances} shows the nonparametric covariances learned by the NP-CGP for each feature in the airline dataset; in our model, we take the product over all of these in order to obtain the full covariance. We can see here that for some features such as \textit{PlaneAge}, \textit{AirTime} and \textit{Distance}, the form of the covariance is very similar to that of the EQ. However, in the covariances for the other features, we see that the NP-CGP has been able to learn a much richer representation of the data, which likely explains the considerable performance gap between the NP-CGP and SGP models on this problem.

\begin{table}[ht!]
    \centering
    \begin{tabular}{ccccc}
    \toprule
    {} & SGP & SGP-500 & DGP2 & NP-CGP \\
    \midrule
     RMSE & 26.2 & 25.9 & 24.9 & \textbf{24.6}  \\
     MNLL & 4.68 & 4.67 & 4.63 & \textbf{4.62} \\
    \bottomrule
    \\
    \end{tabular}
    \caption{Results for the large-scale airline experiment.}
    \label{tab:airline}
\end{table}

\subsection{Deep UCI regression} \label{sec:deep_uci}
To evaluate the performance of our proposed deep model, we replicate the experiments from Section \ref{sec:uci} but for two and three layer versions of our NP-DGP. We compare our model to two and three layer versions of the DGP presented by \citet{salimbeni2017doubly}. From these results, shown in Figure \ref{fig:deep_boxplot}, we can see that the conventional DGPs almost always outperform the NP-DGPs across all four datasets and both evaluation metrics. Additionally, whilst the conventional DGP shows a clear increase in performance as layers are added, the reverse is true for the NP-DGP. Since in theory, our model subsumes the DGP, we attribute the observed difference in performance between the NP-DGP and the DGP to the increased difficulty of the optimisation problem. Further discussion regarding this is provided in Section \ref{sec:discussion}.

\section{Discussion}\label{sec:discussion}
The results of our experiments show that the NP-CGP can provide significant improvements over standard approaches, most notably for large scale regression, however the model does not outperform all of the competing approaches we consider on all problems. The proposed approach seems to exhibit worse performance on the smaller scale problems, and those which have simple structure (i.e. they are well described by linear models). We believe this is due to the mean field variational approximation that is used for the posterior distribution over the covariances. The NP-CGP can over-fit for some problems, learning complex covariances with high confidence, when there is insufficient evidence for such covariances in the data, leading to poor performance on the test set. This does not appear to be a problem for larger datasets, where the model is able to learn covariances much more robustly, leading to improved performance.  \citet{bruinsma2022modelling} recently developed a structured variational scheme for the GPCM, in which the posterior over the input process and convolutional kernel is expressed jointly, allowing for significantly improved quantification of uncertainty for the covariance. The implementation of this improved scheme for the NP-CGP would likely address the over-fitting problem, but is highly non-trivial, therefore we plan to investigate this in future work.

\begin{figure}[t]
    \centering
    \includegraphics[width=1.0\textwidth]{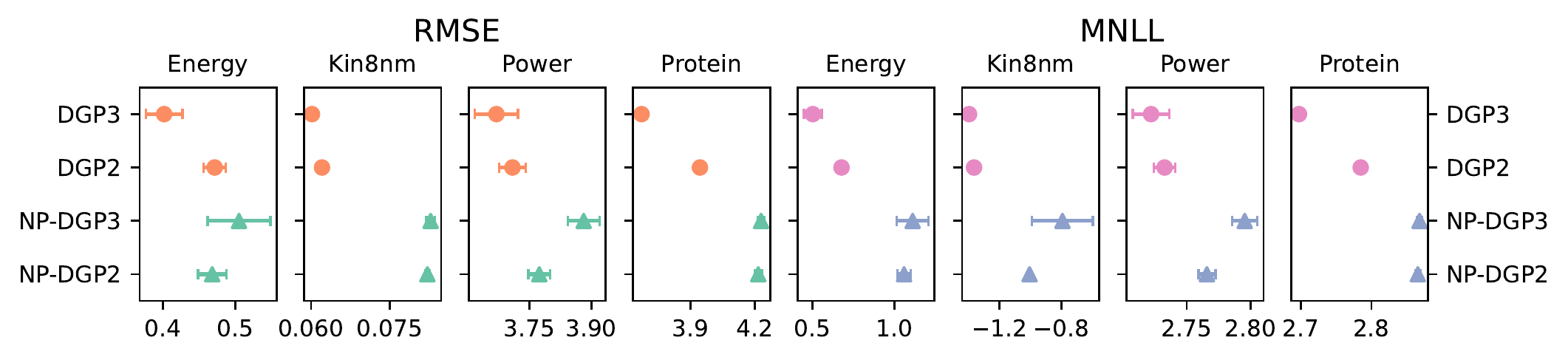}
    \caption{Results over 10 train/test splits for deep models on the UCI experiments.}
    \label{fig:deep_boxplot}
    \vspace{-5mm}
\end{figure}

Whilst the structure of our model easily allows for NP-CGP layers to be composed into a deep architecture, the NP-DGP results presented in Section \ref{sec:deep_uci} show that obtaining competitive performance with such hierarchical models is a challenging task. Whilst the aforementioned issues with the variational inference scheme do not prevent the shallow model from achieving impressive performance in many cases, we posit that these issues are exacerbated in the case of the deep model due to the increased difficulty of the optimisation problem. The shallow model is already capable of representing rich covariance structure, therefore adding further flexibility to the model in the form of additional layers (containing many additional parameters) likely results in a model that cannot be effectively learned using our current mean field inference scheme. This hypothesis is supported by the fact that in Figure \ref{fig:deep_boxplot}, the performance of the three layer NP-DGP is inferior to that of the two layer NP-DGP. This deep architecture does have the capacity to learn extremely expressive covariances, thus we hope that these practical optimisation issues can be tackled as part of future work, be that by the structured variational approach discussed above or by other means.

\section{Conclusion}
In this work we have presented a nonparametric process convolutions model that is suitable for regression tasks with multiple outputs and inputs, along with efficient sampling and inference schemes based on the adaptation of fast functional sampling methods to interdomain GPs. We have shown that allowing the form of the covariance to be directly inferred from the data can lead to increased performance compared to standard GP models across a number of different datasets, in particular for large-scale regression tasks. As we have discussed, the mean field inference scheme we employ has some limitations which particularly affect performance when working with small datasets, or the deep variant of our model; we plan to address these concerns in future work.

\section*{Acknowledgements}
Thomas M. McDonald and Magnus Ross thank the Departments of Computer Science at both the University of Manchester and the University of Sheffield for their financial support, as part of this work was carried out whilst they and Mauricio A. \'Alvarez were at the University of Sheffield. Mauricio A. \'Alvarez has been financed by the EPSRC Research Projects EP/R034303/1, EP/T00343X/2 and EP/V029045/1. The authors would like to acknowledge the assistance given by Research IT and the use of the Computational Shared Facility at The University of Manchester. Additionally, the authors wish to acknowledge CSC – IT Center for Science, Finland, for computational resources.

{
\small
\bibliographystyle{abbrvnat}
\bibliography{refs.bib}
}

\newpage
\appendix

\section{Appendix}
\subsection{Pathwise sampling derivations} \label{sec:samps_app}
To sample from the NP-CGP or DGP-NP, we first must sample from the inducing point variational distributions, $
\mathbf{V}^{\mathbf{G}^\ell} \sim q(\mathbf{V}^{\mathbf{G}^\ell})$ and $
\mathbf{V}^{\tilde{\mathbf{u}}^\ell} \sim q(\mathbf{V}^{\tilde{\mathbf{u}}^\ell})$, then use the pathwise sampling method introduced by \citet{wilson2020efficiently} to sample input functions and convolutional kernel functions for each layer. We then map these analytically through $\mathbf{f}^{(s)}(\mathbf{x}) = \int_{\mathbb{R}^P} \mathbf{G}^{(s)}(\mathbf{x}-\mathbf{z})\mathbf{u}^{(s)}(\mathbf{z})d\mathbf{z}$ in a similar fashion to \citet{ross2021learning}, however in our case, as we have a multi-dimensional input, we split the multi-dimensional integrals involved into products of one dimensional integrals. This process yields a closed form expression for a sample function representing each layer,
\begin{align} \label{dgpnp_sampling}
\begin{split} 
\Big(\mathbf{f}^{{\ell}^{(s)}}|\mathbf{V}^{\mathbf{G}^\ell} , \mathbf{V}^{\tilde{\mathbf{u}}^\ell} \Big)(\mathbf{x}) &= \sum_{k=1}^{N_b} w_k^{u^{\ell}} \Bigg(\frac{e^{i \beta^{u^{\ell}}_k}}{2} \prod^P_{p=1} \bigg( \sum_{i=1}^{N_b} w_i^{G^{\ell}_p} I_{\text{1A}}\Big(x_p ; \alpha^{\ell}_p, \theta_i^{G^{\ell}_p}, \beta_i^{G^{\ell}_p}, \theta_{k, p}^{u^{\ell}} \Big) \\ 
& \qquad \qquad \qquad \qquad \quad + \sum_{j=1}^M q_j^{G^{\ell}_p} I_{\text{1B}}\Big(x_p ; \alpha^{\ell}_p, z_j^{G^{\ell}_p}, \rho^{G^{\ell}_p}, \theta_{k, p}^{u^{\ell}}\Big) \bigg) \\
& \qquad \quad + \frac{e^{-i \beta^{u^{\ell}}_k}}{2} \prod_{p=1}^P \bigg( \sum_{i=1}^{N_b} w_i^{G^{\ell}_p} I_{\text{1A}}\Big(x_p ; \alpha^{\ell}_p, \theta_i^{G^{\ell}_p}, \beta_i^{G^{\ell}_p}, -\theta_{k,p}^{u^{\ell}}\Big) \\
& \qquad \qquad \qquad \qquad \quad + \sum_{j=1}^M q_j^{G^{\ell}_p} I_{\text{1B}}\Big(x_p ; \alpha^{\ell}_p, z_j^{G^{\ell}_p}, \rho^{G^{\ell}_p}, -\theta_{k,p}^{u^{\ell}}\Big) \bigg) \Bigg) \\
& + \sum_{l=1}^M q_l^{u^{\ell}} \prod_{p=1}^P \Bigg( \sum_{i=1}^{N_b} w_i^{G_p^{\ell}} I_{\text{2A}}\Big(x_p ; \alpha_p^{\ell}, \theta_i^{G_p^{\ell}}, \beta_i^{G_p^{\ell}}, \rho_p^{u^{\ell}}, z_{l, p}^{u^{\ell}}\Big) \\
& \qquad \qquad \qquad \quad + \sum_{j=1}^M q_j^{G_p^{\ell}} I_{\text{2B}}\Big(x_p ; \alpha_p^{\ell}, \rho^{G_p^{\ell}}, z_j^{G_p^{\ell}}, \rho_p^{u^{\ell}}, z_{l, p}^{u^{\ell}}\Big) \Bigg),
\end{split}
\end{align}
where,
\begin{align}
\begin{split}
    I_{\text{1A}} & \Big(x ; \alpha, \theta_1, \beta, \theta_2 \Big) = \int_{-\infty}^{\infty} e^{-\alpha (x - \tau)^2} \cos\left(\theta_1 \tau + \beta\right) e^{i\theta_2\tau} d\tau \\
    & \qquad = \frac{\sqrt{\pi}}{2\sqrt{\alpha}} \left( \exp\left(\frac{\theta_1 \theta_2}{\alpha}\right) + \exp \left( 2i(\theta_1 x + \beta) \right) \right) \times \\ & \qquad \qquad \qquad \exp \left(-i(\beta + (\theta_1 - \theta_2)x) - \frac{-(\theta_1 + \theta_2)^2}{4\alpha} \right) ,
\end{split}
\end{align}

\begin{align}
\begin{split}
    I_{\text{1B}} & \Big(x ; \alpha, z, \rho, -\theta \Big) = \int_{-\infty}^{\infty} e^{-\alpha (x - \tau)^2} k(\tau, z)  e^{i\theta\tau} d\tau \\
    & \qquad \qquad = \frac{\sqrt{\pi}}{\sqrt{\alpha + \rho}} \exp \left(- \frac{\theta^2 + 4\alpha \rho(x-z)^2 - 4i\theta(\alpha x + \rho z)}{4(\alpha + \rho)} \right) ,
\end{split}
\end{align}

\begin{align}
\begin{split}
    I_{\text{2A}} & \Big(x ; \alpha, \theta, \beta, \rho, z\Big) = \int_{-\infty}^{\infty} e^{-\alpha (x - \tau)^2} \cos\left(\theta \tau + \beta \right) e^{-\rho (\tau - z)^2} d\tau \\
    & \qquad = \frac{\sqrt{\pi}}{\sqrt{\alpha + \rho}} \cos \left( \beta + \theta \left(x - \frac{\rho z}{\sqrt{\alpha + \rho}} \right) \right) \exp \left({-\frac{(\theta)^2 + 4 \alpha }{ \rho (z)^2} } \right),
\end{split}
\end{align}

\begin{align}
\begin{split}
    I_{\text{2B}} & \Big(x ; \alpha, \rho_1, z_1, \rho_2, z_2\Big) = \int_{-\infty}^{\infty} e^{-\alpha (x - \tau)^2} k(\tau, z_2) e^{-\rho_2 (\tau - z_2)^2} d\tau \\
    & \qquad = \frac{\sqrt{\pi}}{\sqrt{\alpha + \rho_1 + \rho_2}} \exp \left( { \frac{ \alpha \left( \rho_1  (z_1)^2 + \rho_2 (x - z_2)^2 \right) + \rho_1 \rho_2 \left(-x + z_1 + z_2 \right)^2 }{\alpha + \rho_1 + \rho_2} } \right).
\end{split}
\end{align}
We can then compute the output samples from the NP-DGP as, 
\begin{equation}
        \mathbf{F}^{{L}^{(s)}} = \Big(\mathbf{f}^L|\mathbf{V}^{\mathbf{G}^L} , \mathbf{V}^{\tilde{\mathbf{u}}^L} \Big)\circ \Big(\mathbf{f}^{{L-1}^{(s)}}|\mathbf{V}^{\mathbf{G}^{L-1}} , \mathbf{V}^{\tilde{\mathbf{u}}^{L-1}} \Big) \circ \dots \circ \Big(\mathbf{f}^{1^{(s)}}|\mathbf{V}^{\mathbf{G}^1} , \mathbf{V}^{\tilde{\mathbf{u}}^1} \Big)(\mathbf{x}). 
\end{equation}

\subsection{Interdomain sampling derivations}
In the main paper, we use the closed form expression for the Gaussian transformed basis functions in order to sample from the interdomain input process in our model. For an input observation $\mathbf{x} \in \mathbb{R}^P$, and a given latent process $\tilde{u}_q(\mathbf{x})$ (we omit the subscript for ease of exposition below), this expression takes the form,
\begin{align}
\begin{split}
    \tilde{u}(\mathbf{x}) &= \sum_{j=1}^{N_b} w_j \int_{\mathbb{R}^P} e^{-\sum_{i=1}^{P}\rho_i (z_i - x_i)^2 } \cos\left(\boldsymbol{\theta}_j^\top \mathbf{z} + \beta_j \right) d\mathbf{z} \\
    &= \sum_{j=1}^{N_b} \frac{e^{-\beta_j}}{2} \prod_{i=1}^P \int_{-\infty}^{\infty} e^{-\rho_i (z - x_i)^2 - i\theta_{i, j} z} dz + \frac{e^{\beta_j}}{2} \prod_{i=1}^P \int_{-\infty}^{\infty} e^{-\rho_i (z - x_i)^2 + \theta_{i, j} z} dz,
\end{split}
\end{align}
where $\mathbf{w}$, $\boldsymbol{\beta}$ and $\boldsymbol{\theta}$ define our random Fourier feature basis. $\mathbf{w}$ consists of entries $w_j \sim \mathcal{N}(0, 1)$, $\boldsymbol{\beta}$ consists of entries $\beta_j \sim U(0, 2\pi)$ and $\boldsymbol{\theta}$ consists of entries $\boldsymbol{\theta}_j \sim \text{FT}(k)$, where $\text{FT}$ is the Fourier transform and $k$ represents the covariance of the untransformed process. This transform allows us to compute the elements of $\tilde{\boldsymbol{\Phi}}$ in Eq. 5 of the main paper, which in turn allows us to sample from our input process. Eq. 5 itself follows from a reformulation of Eq. 13 in \citet{wilson2020efficiently},
\begin{align}
(u_q^{(s)} \mid \mathbf{v}^{u_q})(\cdot) = \sum_{i=1}^{N_b} w_{i} \phi_{i}(\cdot) + k\left(\cdot, \boldsymbol{z}^{u_q}\right) \mathbf{K}_{u_q, u_q}^{-1}(\mathbf{v}^{u_q} - \boldsymbol{\Phi}\mathbf{w}) .
\end{align}
In this work, as we are using an interdomain input process, we must ensure that the update term in this expression is computed in the transformed domain. This is achieved by replacing the cross-covariance $k\left(\cdot, \boldsymbol{z}^{u_q}\right)$ with $k_{u_q, \tilde{u}_q}\left(\cdot, \boldsymbol{z}^{\tilde{u}_q}\right)$ and the covariance matrix $\mathbf{K}_{u_q, u_q}^{-1}$ with its interdomain equivalent $\mathbf{K}_{\tilde{u}_q, \tilde{u}_q}^{-1}$. We also replace the inducing points $\mathbf{v}^{u_q}$ with the interdomain inducing points $\mathbf{v}^{\tilde{u}_q}$, and the basis functions $\boldsymbol{\Phi}$ with their transformed counterpart $\tilde{\boldsymbol{\Phi}}$. Applying these changes yields Eq. 5, as stated in the main paper. The covariance used to compute the elements of $\mathbf{K}_{\tilde{u}_q, \tilde{u}_q}$ can be expressed as,
\begin{equation}
    k_{\tilde{u}_q, \tilde{u}_q}\left(\mathbf{z}^{\tilde{u}_q}, {\mathbf{z}^{\tilde{u}_q}}^\prime \right) = \int\int_{\mathbb{R}^P} k_{u_q}(\mathbf{x}, \mathbf{x}^\prime) g_q(\mathbf{x}, \mathbf{z}^{\tilde{u}_q}) g_q(\mathbf{x}^\prime, {\mathbf{z}^{\tilde{u}_q}}^\prime) d\mathbf{x}^\prime d\mathbf{x} ,
\end{equation}
whilst the aforementioned cross-covariance can be expressed as,
\begin{equation}
    k_{u_q, \tilde{u}_q}\left(\mathbf{x}, {\mathbf{z}^{\tilde{u}_q}}^\prime \right) = \int_{\mathbb{R}^P} k_{u_q}(\mathbf{x}, \mathbf{x}^\prime) g_q(\mathbf{x}^\prime, {\mathbf{z}^{\tilde{u}_q}}^\prime) d\mathbf{x}^\prime .
\end{equation}

\subsection{Variational lower bound derivation} \label{sec:elbo_app}
In this section, we present the full derivation of the variational lower bound for the NP-DGP. Denoting our input data as $\mathbf{X}\in\mathbb{R}^{N\times P}$, and the corresponding outputs as $\mathbf{Y}\in\mathbb{R}^{N\times D_{L}}$, we can express the joint distribution of the DGP-NP as,
\begin{equation} \label{dgpnp_joint}
    p(\mathbf{Y}, \{\mathbf{G}^\ell, \mathbf{V}^{\mathbf{G}^\ell} , \mathbf{u}^\ell, \mathbf{V}^{\tilde{\mathbf{u}}^\ell} \}^L_{\ell = 1}) =\prod^{N}_{i=1}  p(\mathbf{y}_{i}|\mathbf{F}_i^L)
\prod^{L}_{\ell=1} p(\mathbf{G}^\ell | \mathbf{V}^{\mathbf{G}^\ell})p(\mathbf{V}^{\mathbf{G}^\ell})p(\mathbf{u}^\ell | \mathbf{V}^{\tilde{\mathbf{u}}^\ell}) p(\mathbf{V}^{\tilde{\mathbf{u}}^\ell}) ,
\end{equation}
where the likelihood is given by $p(\mathbf{y}_{i}|\mathbf{F}_i^L) = \mathcal{N}(\mathbf{y}_i ; \mathbf{F}_i^L, \sigma^2_Y)$. As all of the convolutional kernel and input GPs are independent, we have $p(\mathbf{G}^\ell | \mathbf{V}^{\mathbf{G}^\ell}) = \prod^{D_{\ell}}_{d=1} p( G^{\ell}_{d}| \mathbf{v}^{G^{\ell}_{d}})$,  where $p( G^{\ell}_{d}| \mathbf{v}^{G^{\ell}_{d}})$ is the GP posterior distribution given the inducing points, and likewise $p(\mathbf{u}^\ell | \mathbf{V}^{\tilde{\mathbf{u}}^\ell}) = \prod^{Q_\ell}_{q=1} p( u^{\ell}_{ q}| \mathbf{v}^{\tilde{u}^{\ell}_{q}})$, where again $p( u^{\ell}_{q}| \mathbf{v}^{\tilde{u}^{\ell}_{q}})$ are GP posteriors. $p(\mathbf{V}^{\mathbf{G}^\ell})$ and $p(\mathbf{V}^{\tilde{\mathbf{u}}^\ell})$ represent the priors over the inducing points. As mentioned in the main text, whilst here we discuss our inference procedure in the context of the DGP-NP, the corresponding expressions for the NP-CGP and FNP-CGP can be recovered by setting $L=1$. Our variational posterior takes the form,
\begin{equation} \label{dgpnp_posterior}
    q(\{\mathbf{G}^\ell, \mathbf{V}^{\mathbf{G}^\ell} , \mathbf{u}^\ell, \mathbf{V}^{\tilde{\mathbf{u}}^\ell} \}^L_{\ell = 1}) = \prod^L_{\ell = 1} p(\mathbf{G}^\ell | \mathbf{V}^{\mathbf{G}^\ell})q(\mathbf{V}^{\mathbf{G}^\ell})p(\mathbf{u}^\ell | \mathbf{V}^{\tilde{\mathbf{u}}^\ell})q(\mathbf{V}^{\tilde{\mathbf{u}}^\ell}),
\end{equation}
where $q(\mathbf{V}^{\mathbf{G}^\ell}) =\prod^{D_{\ell}}_{d=1} \mathcal{N}(\mathbf{v}^{G^{\ell}_{d}};\boldsymbol{\mu}^{G^{\ell}_{d}}, \boldsymbol{\Sigma}^{G^{\ell}_{d}})$  and $q(\mathbf{V}^{\tilde{\mathbf{u}}^\ell}) =\prod^{Q_{\ell}}_{ q=1} \mathcal{N}(\mathbf{v}^{\tilde{u}^{\ell}_{q}};\boldsymbol{\mu}^{\tilde{u}^{\ell}_{q}}, \boldsymbol{\Sigma}^{\tilde{u}^{\ell}_{q}})$ are layer-specific variational distributions, whose means and covariance matrices are variational parameters. For ease of exposition, we have omitted the factorisation of the posterior over the layer dimensionality. We can write down the variational lower bound as,
\begin{align}
    \mathcal{L} &= \mathbb{E}_{q(\{\mathbf{G}^\ell, \mathbf{V}^{\mathbf{G}^\ell} , \mathbf{u}^\ell, \mathbf{V}^{\tilde{\mathbf{u}}^\ell} \}^L_{\ell = 1})} \left[\frac{p(\mathbf{Y}, \{\mathbf{G}^\ell, \mathbf{V}^{\mathbf{G}^\ell} , \mathbf{u}^\ell, \mathbf{V}^{\tilde{\mathbf{u}}^\ell} \}^L_{\ell = 1})}{q(\{\mathbf{G}^\ell, \mathbf{V}^{\mathbf{G}^\ell} , \mathbf{u}^\ell, \mathbf{V}^{\tilde{\mathbf{u}}^\ell} \}^L_{\ell = 1})} \right] \\
    &= \int q(\{\mathbf{G}^\ell, \mathbf{V}^{\mathbf{G}^\ell} , \mathbf{u}^\ell, \mathbf{V}^{\tilde{\mathbf{u}}^\ell} \}^L_{\ell = 1}) \log \left[\frac{p(\mathbf{Y}, \{\mathbf{G}^\ell, \mathbf{V}^{\mathbf{G}^\ell} , \mathbf{u}^\ell, \mathbf{V}^{\tilde{\mathbf{u}}^\ell} \}^L_{\ell = 1})}{q(\{\mathbf{G}^\ell, \mathbf{V}^{\mathbf{G}^\ell} , \mathbf{u}^\ell, \mathbf{V}^{\tilde{\mathbf{u}}^\ell} \}^L_{\ell = 1})} \right] d\mathbf{S} ,
\end{align}
where $d\mathbf{S}$ represents the integral over all of the inducing points, convolutional kernel and input processes, across all layers of the model. Using Eq. \ref{dgpnp_joint} and Eq. \ref{dgpnp_posterior}, we can derive a form of the evidence lower bound (ELBO) which we can use to perform approximate inference as follows:
\begin{align}
    \begin{split}
        \mathcal{L} &= \int \prod^L_{\ell = 1} p(\mathbf{G}^\ell | \mathbf{V}^{\mathbf{G}^\ell})q(\mathbf{V}^{\mathbf{G}^\ell})p(\mathbf{u}^\ell | \mathbf{V}^{\tilde{\mathbf{u}}^\ell})q(\mathbf{V}^{\tilde{\mathbf{u}}^\ell}) \\
        & \quad \quad \times \log \left[\frac{\prod^{N}_{i=1}  p(\mathbf{y}_{i}|\mathbf{F}_i^L)
        \prod^{L}_{\ell=1} \cancel{p(\mathbf{G}^\ell | \mathbf{V}^{\mathbf{G}^\ell})}p(\mathbf{V}^{\mathbf{G}^\ell})\cancel{p(\mathbf{u}^\ell | \mathbf{V}^{\tilde{\mathbf{u}}^\ell})} p(\mathbf{V}^{\tilde{\mathbf{u}}^\ell})}{\prod^L_{\ell = 1} \cancel{p(\mathbf{G}^\ell | \mathbf{V}^{\mathbf{G}^\ell})}q(\mathbf{V}^{\mathbf{G}^\ell})\cancel{p(\mathbf{u}^\ell | \mathbf{V}^{\tilde{\mathbf{u}}^\ell})}q(\mathbf{V}^{\tilde{\mathbf{u}}^\ell})} \right] d\mathbf{S}
    \end{split} \\
    \begin{split}
        &= \int \prod^L_{\ell = 1} p(\mathbf{G}^\ell | \mathbf{V}^{\mathbf{G}^\ell})q(\mathbf{V}^{\mathbf{G}^\ell})p(\mathbf{u}^\ell | \mathbf{V}^{\tilde{\mathbf{u}}^\ell})q(\mathbf{V}^{\tilde{\mathbf{u}}^\ell}) \\
        & \quad \quad \times \log \left[\frac{\prod^{N}_{i=1}  p(\mathbf{y}_{i}|\mathbf{F}_i^L)
        \prod^{L}_{\ell=1} p(\mathbf{V}^{\mathbf{G}^\ell}) p(\mathbf{V}^{\tilde{\mathbf{u}}^\ell})}{\prod^L_{\ell = 1} q(\mathbf{V}^{\mathbf{G}^\ell})q(\mathbf{V}^{\tilde{\mathbf{u}}^\ell})} \right] d\mathbf{S}
    \end{split} \\
    \begin{split}
        &= \int \prod^L_{\ell = 1} p(\mathbf{G}^\ell | \mathbf{V}^{\mathbf{G}^\ell})q(\mathbf{V}^{\mathbf{G}^\ell})p(\mathbf{u}^\ell | \mathbf{V}^{\tilde{\mathbf{u}}^\ell})q(\mathbf{V}^{\tilde{\mathbf{u}}^\ell}) \log \left[\prod^{N}_{i=1}  p(\mathbf{y}_{i}|\mathbf{F}_i^L)\right] d\mathbf{S} \\
        & \quad - \sum^L_{\ell = 1} \left[ \text{KL}[q(\mathbf{V}^{\tilde{\mathbf{u}}^\ell}) \lVert p(\mathbf{V}^{\tilde{\mathbf{u}}^\ell})] + \text{KL}[q(\mathbf{V}^{\mathbf{G}^\ell}) \lVert p(\mathbf{V}^{\mathbf{G}^\ell})] \right]
    \end{split} \\
    \begin{split} \label{dgpnp_elbo}
        &= \sum^N_{i=1} \mathbb{E}_{q(\{\mathbf{G}^\ell, \mathbf{V}^{\mathbf{G}^\ell} , \mathbf{u}^\ell, \mathbf{V}^{\tilde{\mathbf{u}}^\ell} \}^L_{\ell = 1})} \left[\log p(\mathbf{y}_{i}|\mathbf{F}_i^L) \right] \\
        & \quad - \sum^L_{\ell = 1} \left[ \text{KL}[q(\mathbf{V}^{\tilde{\mathbf{u}}^\ell}) \lVert p(\mathbf{V}^{\tilde{\mathbf{u}}^\ell})] + \text{KL}[q(\mathbf{V}^{\mathbf{G}^\ell}) \lVert p(\mathbf{V}^{\mathbf{G}^\ell})] \right]
    \end{split}
\end{align}
As the KL divergences present are between sets of independent multivariate Gaussian distributions, we have $\text{KL}[q(\mathbf{V}^{\tilde{\mathbf{u}}^\ell}) \lVert p(\mathbf{V}^{\tilde{\mathbf{u}}^\ell})] = \sum^Q_{q=1}\text{KL}[\mathcal{N}(\mathbf{v}^{\tilde{u}^{\ell}_{q}};\boldsymbol{\mu}^{\tilde{u}^{\ell}_{q}}, \boldsymbol{\Sigma}^{\tilde{u}^{\ell}_{q}})\lVert\mathcal{N}(\mathbf{v}^{\tilde{u}^{\ell}_{q}};0, \boldsymbol{K}^{\tilde{u}^{\ell}_{q}})]$ and $\text{KL}[q(\mathbf{V}^{\mathbf{G}^\ell}) \lVert p(\mathbf{V}^{\mathbf{G}^\ell})] = \sum^{D}_{d=1}\text{KL}[\mathcal{N}(\mathbf{v}^{G^{\ell}_{d}};\boldsymbol{\mu}^{G^{\ell}_{d}}, \boldsymbol{\Sigma}^{G^{\ell}_{d}})\lVert\mathcal{N}(\mathbf{v}^{G^{\ell}_{d}};0, \boldsymbol{K}^{G^{\ell}_{d}})]$, which have well known tractable form. Conversely, we approximate the intractable expectation from the first line of Eq. \eqref{dgpnp_elbo} stochastically using $S$ Monte Carlo samples,
\begin{equation} \label{dgpnp_ell}
    \mathbb{E}_{q(\{\mathbf{G}^\ell, \mathbf{V}^{\mathbf{G}^\ell} , \mathbf{u}^\ell, \mathbf{V}^{\tilde{\mathbf{u}}^\ell} \}^L_{\ell = 1})} \left[\log p(\mathbf{y}_{i}|\mathbf{F}_i^L) \right] \approx \frac{1}{S} \sum^S_{s=1} \log p(\mathbf{y}_i | \mathbf{F}_i^{L^{(s)}}).
\end{equation}
where $\mathbf{F}_i^{L^{(s)}}$ denotes a sample from the model.

\section{Model complexities and runtimes}
Table \ref{tab:model_complexities} shows complexities for the computation of the bound/sampling for the regular version of our shallow NP-CGP model and the fast approximation, for the cases of single and multiple outputs. We present these alongside the complexities associated with comparable prior work. Note that the complexities do not depend on the size of the data because we employ mini-batching. Recall that $M_u$ is the number of input process inducing points, $M_G$ the number of convolutional kernel process inducing points, $M$ is the number of regular GP inducing points, $P$ denotes the input dimension, $D$ denotes the output dimension, and $Q$ denotes the number of latent functions. 
\begin{table}[!ht]
    \centering
    \begin{tabular}{cccc}
    \toprule
{} &  & \multicolumn{1}{c}{Model Type} \\
\cmidrule{2-4}

        {} & Regular & Fast & Prior work \\
        \midrule
        Single output & $\mathcal{O}(M_u^3+PM_G^3)$ & $\mathcal{O}(M_u^3+PM_G^3)$ & $\mathcal{O}(M^3)$ (SGP) \\ 
        Multi output & $\mathcal{O}(QM_u^3+PDM_G^3)$ & $\mathcal{O}(QM_u^3+PM_G^3)$ & $\mathcal{O}(DM^3)$(SLMC) \\
        \bottomrule\\
    \end{tabular}
        \caption{Computational complexities for the standard and fast versions of the model, and the equivalent prior work.}
    \label{tab:model_complexities}
\end{table}

\par 
In practice this corresponds to the run-time of our model being around $15\times$ longer than a standard SGP model. Although this seems drastically slower, this is not a fundamental limitation of the model, and is due to the somewhat complex implementation of our models. We believe the code could likely be optimised in such a way that the model could be made to run much faster. In particular, the computation of gradients with respect to the integrals required for sampling took significantly longer than expected, as the PyTorch framework is not well optimised for computing gradients with respect to the complex point-wise operations represented by the integrals in the model.

\subsection{Experimental details}
The experiments in this work were performed on HPC clusters, using nodes containing a range of different GPUs, including 40GB and 80GB NVIDIA A100-SXM4 GPUs, as well as 16GB and 32GB NVIDIA Tesla V100-SXM2 GPUs. Throughout we utilise the interdomain transform for the input process, since without this element we were unable to achieve convergence during training for most problems. 

\subsection{Toy experiment}
For our toy experiment, we used an input $\mathbf{x} \in \mathbb{R}^{N \times P}$, with $N=3000$ and $P=2$, where the entries of $\mathbf{x}$ were sampled from a standard normal distribution. As mentioned in the main paper, the ground truth function values in this experiment, $f_1(\mathbf{x})$ and $f_2(\mathbf{x})$, were generated by sampling from two different linear combinations of two GP priors $u_{\text{EQ}}\sim \mathcal{GP}(0, k_{\text{EQ}})$ and $u_{\text{P}}\sim \mathcal{GP}(0, k_{\text{P}})$, which have EQ and weakly periodic kernels respectively. $k_{\text{EQ}}$ uses a lengthscale of 1.5 for each input dimension, and $k_{\text{P}}$ is constructed from the product of an EQ kernel and periodic kernel both using lengthscales of 1.5 per dimension, with periods of 1.8 and 2.1 for each input dimension in the periodic component. Specifically, the linear combinations we use are $f_1 = 0.9u_{\text{EQ}} + 0.1u_P$ and $f_2 = 0.5u_{\text{EQ}} + 0.5u_P$. Prior to sampling our ground truth output values, we also applied independent Gaussian noise to each output with $\sigma = 0.01$.

\begin{table*}
    \centering
    \begin{tabular}{cccccccc}
    \toprule
    {} &  &  & \multicolumn{5}{c}{RMSE} \\
    \cmidrule{4-8}
    {} & N & P & NP-CGP & NP-CGP-300 & SGP & SGP-300 & SGP-M32 \\
    \midrule
    energy  & 768 & 8 &  0.52 (0.03) &  0.81 (0.04) &  0.46 (0.01) &  0.46 (0.01) & 0.45 (0.01) \\
    kin8nm  & 8192 & 8 & 0.08 (0.00) &  0.09 (0.00) &  0.08 (0.00) &  0.08 (0.00) &  0.09 (0.00) \\
    power   & 9568 & 4 & 3.82 (0.03) & 3.76 (0.03) &  3.85 (0.03) & 3.73 (0.03) &  3.77 (0.03) \\
    protein & 45730 & 9 & 4.22 (0.01) & 3.98 (0.01) &  4.41 (0.01) &  4.16 (0.01) &  4.31 (0.01) \\
    \bottomrule
    \\
    \end{tabular}   
    
    \begin{tabular}{cccccccc}
    \toprule
    {} & & & \multicolumn{5}{c}{MNLL} \\
    \cmidrule{4-8}
    {} & N & P &  NP-CGP & NP-CGP-300 &  SGP &  SGP-300 & SGP-M32 \\
    \midrule
    energy  & 768 & 8 & \ 1.08 (0.07) &   \ 1.67 (0.08) &  \ 0.65 (0.03) &   \ 0.64 (0.03) &  \ 0.64 (0.03) \\
    kin8nm  & 8192 & 8 &  -1.02 (0.01) &  -0.37 (0.03) & -1.03 (0.00) & -1.13 (0.00) &  -0.98 (0.00) \\
    power   & 9568 & 4 &  \ 2.78 (0.01) &  \ 2.77 (0.01) & \ 2.77 (0.01) &  \ 2.74 (0.01) & \ 2.75 (0.01) \\
    protein & 45730 & 9 & \ 2.86 (0.00) &  \ 2.81 (0.00) & \ 2.90 (0.00) & \ 2.84 (0.00) &  \ 2.88 (0.00) \\
    \bottomrule
    \\
    \end{tabular}
    \caption{Results over 20 train/test splits for the UCI regression experiments. The mean values are reported, with the standard error in brackets. N represents the number of observations in each dataset, and P the number of input dimensions.}
    \label{tab:uci_results_shallow}
\end{table*}

\subsection{UCI regression} \label{sec:uci_shallow_app}
The numerical values used to generate the boxplots in Figure 4 in the main paper, are shown in Table \ref{tab:uci_results_shallow}. As discussed in the main paper, all data is freely available from the UCI Machine Learning Repository \citep{Dua_2019}. For these experiments, we performed 20 different random splits of the standardised data (which were kept in common across all of the models evaluated), using 90\% of each dataset for training and the remaining 10\% to evaluate the test set metrics which we report in the paper and this appendix. For our NP-CGP models, we used 16 basis functions and $S=2$ Monte Carlo samples for estimating the lower bound, initialising the likelihood variance to 0.01. We used 15 inducing points for the convolutional kernel processes, and either 100 (NP-CGP) or 300 (NP-CGP-300) inducing points for the input processes, and these input inducing points were initialised using $k$-means clustering. Training was performed for 40,000 iterations with a batch size of 1000, using the \textit{Adam} optimiser \citep{DBLP:journals/corr/KingmaB14} with a learning rate of $0.001$. For the variational GP models (SGP, SGP-300 and SGP-M32), we mirrored these settings as closely as possible, with the only difference being that we used a learning rate of 0.01 for the optimiser.

\begin{table*}
\centering
\begin{tabular}{cccccccc}
\toprule
{} & & & & \multicolumn{4}{c}{RMSE} \\
\cmidrule{5-8}
{} & N & P & D & FNP-CGP & NP-CGP & S-LMC & CMOGP  \\
\midrule
energy  & 768 & 8 & 2 & 1.19 (0.02) &  0.84 (0.03) &  1.27 (0.00) &  0.89 (0.08) \\
naval   & 11934 & 16 & 2 & 0.00 (0.00) &  0.00 (0.00) &  0.00 (0.00) &  \qquad  - \\
polymer & 60 & 10 & 4 & 0.08 (0.01) &  0.08 (0.00) &  0.12 (0.00) &  0.19 (0.03) \\
\bottomrule
\\
\end{tabular}

\begin{tabular}{cccccccc}
\toprule
{} & & & & \multicolumn{4}{c}{MNLL} \\
\cmidrule{5-8}
{} & N & P & D & FNP-CGP &  NP-CGP & S-LMC & CMOGP \\
\midrule
energy  & 768 & 8 & 2 &  \ 1.60 (0.03) &   \ 1.46 (0.04) &  \ 2.16 (0.01) &   \ 1.44 (0.23) \\
naval  & 11934 & 16 & 2 & -5.38 (0.06) &  -5.21 (0.04) &  -7.55 (0.01) &    \qquad - \\
polymer & 60 & 10 & 4 & \ 0.63 (0.91) &  -0.77 (0.04) &   \ 0.00 (0.00) &  -0.15 (0.16) \\
\bottomrule
\\
\end{tabular}
    \caption{Results over 20 train/test splits for the experiments with multiple inputs and outputs. The mean values are reported, with the standard error in brackets. N represents the number of observations in each dataset, P the number of input dimensions and D the number of outputs.}
    \label{tab:mo_results}
\end{table*}

\subsection{Regression with multiple inputs and outputs}
The numerical values used to generate the boxplots in Figure 5 in the main paper, are shown in Table \ref{tab:mo_results}.
We include results for regression with multiple inputs and outputs on three different datasets. Firstly, the \textit{energy} dataset is the same dataset used in the  UCI experiments, however rather than just using one of the two outputs, we infer both. Similarly, \textit{naval} is another UCI dataset commonly used as a single output benchmark, but we infer both of its outputs in this work. Finally, the \textit{polymer} dataset, freely available at \url{ftp://ftp.cis.upenn.edu/pub/ungar/ chemdata}, was selected in order to test the predictive capability of the model in a small-data setting. The settings used for our NP-CGP models in this experiment broadly mirror those described in Section \ref{sec:uci_shallow_app}. For the stochastic multi-output GP (S-LMC), implemented using GPyTorch \citep{gardner2018gpytorch}, we used a number of latent GPs equal to the number of outputs for the given dataset, and all experimental settings for this model were the same as those used for the NP-CGP. Similarly, for the convolved MOGP (CMOGP), we also used a number of latent functions equal to the number of outputs.

\subsection{Large-scale regression}
The settings used for the NP-CGP on the large-scale regression experiment once again broadly mirror those described in Section \ref{sec:uci_shallow_app}, with three key differences due to the scale of the dataset: firstly the batch size was increased to 10,000, secondly, the number of training iterations was increased to 100,000, and finally, we did not perform repeats. For the DGP2 model we employ EQ ARD kernels and the same initialisations used by \citet{salimbeni2017doubly}.

\subsection{Deep UCI regression}
The numerical values used to generate the boxplots in Figure 7 in the main paper, are shown in Table \ref{tab:uci_results_deep}. For the NP-DGP models we tested, the experimental settings used again closely follow those specified in Section \ref{sec:uci_shallow_app}, except that now we extend that model to multiple layers. For the regular DGP models, we followed the procedure of \citet{salimbeni2017doubly} as closely as possible in order to replicate their results.

\begin{table*}
    \centering
\begin{tabular}{ccccccc}
\toprule
{} & & & \multicolumn{4}{c}{RMSE} \\
\cmidrule{4-7}
{} & N & P & NP-DGP2 & NP-DGP3 &  DGP2 & DGP3 \\
\midrule
energy  & 768 & 8 & 0.47 (0.03) &  0.50 (0.04) & 0.47 (0.02) &  0.40 (0.03) \\
kin8nm & 8192 & 8 & 0.08 (0.00) &  0.08 (0.00) &  0.06 (0.00) &  0.06 (0.00) \\
power & 9568 & 4 & 3.77 (0.04) &  3.88 (0.04) & 3.71 (0.05) & 3.67 (0.05) \\
protein & 45730 & 9 & 4.21 (0.02) &  4.23 (0.01) & 3.94 (0.01) & 3.67 (0.01) \\
\bottomrule
\\
\end{tabular}    

\begin{tabular}{ccccccc}
\toprule
{} & & & \multicolumn{4}{c}{MNLL} \\
\cmidrule{4-7}
{} & N & P & NP-DGP2 & NP-DGP3 & DGP2 & DGP3 \\
\midrule
energy  & 768 & 8 & \ 1.06 (0.06) &   \ 1.11 (0.10) &   \ 0.68 (0.05) &   \ 0.50 (0.06) \\
kin8nm  & 8192 & 8 & -1.00 (0.01) &  -0.79 (0.20) &  -1.36 (0.00) &  -1.40 (0.00) \\
power  & 9568 & 4 & \ 2.77 (0.01) &  \ 2.80 (0.01) &   \ 2.73 (0.01) &   2.72 (0.01) \\
protein & 45730 & 9 & \ 2.87 (0.00) &   \ 2.87 (0.00) &   \ 2.78 (0.00) &   \ 2.70 (0.00) \\
\bottomrule
\\
\end{tabular}
    \caption{Results over 10 train/test splits for the deep UCI regression experiments. The mean values are reported, with the standard error in brackets. The number of observations and input dimensionality for each of these datasets is shown in Table \ref{tab:uci_results_shallow}.}
    \label{tab:uci_results_deep}
\end{table*}

\end{document}